\documentclass[11pt]{article}

\usepackage[final]{acl}

\usepackage{times}
\usepackage{latexsym}

\usepackage[T1]{fontenc}
\usepackage[frozencache,cachedir=minted-cache]{minted}
\usepackage{newunicodechar}
\newunicodechar{≤}{$\leq$}
\newunicodechar{≥}{$\geq$}
\newunicodechar{≠}{$\neq$}
\newunicodechar{Σ}{$\sum$}
\newunicodechar{√}{$\sqrt{}$}

\usepackage[utf8]{inputenc}

\usepackage{microtype}

\usepackage{inconsolata}

\usepackage{graphicx}
\usepackage{svg}

\usepackage{amsmath}
\usepackage{amsfonts}
\usepackage{algorithm}
\usepackage{algorithmic}
\usepackage{booktabs}
\usepackage{multirow}
\usepackage{enumitem}
\usepackage{pifont}% http://ctan.org/pkg/pifont

\usepackage{listings}
\usepackage{xcolor}

\newcommand{\NULL}{\texttt{NULL}}

\newcommand{\method}{\textsc{CodeEvolve}}
\newcommand{\minmax}{\texttt{Min\-i\-mize\-Max\-Min\-Dist}}
\newcommand{\circlepacking}{\texttt{Cir\-cle\-Pack\-ing\-Square}}

\newcommand{\mainrepourl}{https://github.com/inter-co/science-codeevolve}
\newcommand{\codeevolvemainrepo}{our project repository}

\title{\method{}: an open-source evolutionary framework for\\algorithmic discovery and optimization}

\author{
 \textbf{Henrique Assumpção\textsuperscript{1,2}},
 \textbf{Diego Ferreira\textsuperscript{1,2}},
 \textbf{Leandro Campos\textsuperscript{1,2}},
 \textbf{Fabricio Murai\textsuperscript{3}}
\\
 \textsuperscript{1}Inter Science - Inter\&Co, Belo Horizonte, MG, Brazil \\
 \textsuperscript{2}Federal University of Minas Gerais, Belo Horizonte, MG, Brazil \\
 \textsuperscript{3}Worcester Polytechnic Institute, Worcester, MA, USA
\\
 \small{
   \textbf{Correspondence:} \href{mailto:henrique.soares@inter.co}{henrique.soares@inter.co}
 }
}

\begin{document}
\maketitle

\begin{abstract}
We introduce \method{}, an open-source framework that couples large language models with island-based evolutionary search for end-to-end algorithmic discovery. \method{} integrates inspiration-based crossover, meta-prompting, and depth-based refinement on top of a CVT-MAP-Elites archive and a weighted LLM ensemble to generate optimized solutions for complex problems. On the AlphaEvolve benchmark suite, \method{} matches or surpasses the reported AlphaEvolve results on 5 of 9 problems and, under matched conditions, outperforms the open-source frameworks OpenEvolve and ShinkaEvolve on 6 of 9. With the open-weight \texttt{Qwen3-Coder-30B} backbone, it surpasses the reported AlphaEvolve score on both \texttt{CirclePackingSquare} instances at roughly an order of magnitude lower cost than a frontier closed-source ensemble, and remains competitive with EoH on heuristic-design tasks without retuning. Ablations show that the interaction between \method{}'s components, rather than any single operator, drives these results. We release the framework, experimental data, and practical hyperparameter guidelines at \codeevolvemainrepo.\footnote{\url{\mainrepourl}}
\end{abstract}

\section{Introduction}

The intersection of large language models (LLMs) and algorithmic reasoning has catalyzed a shift from routine program synthesis to automated scientific discovery. While early work demonstrated the ability of transformers to synthesize code from natural language~\citep{chen2021evaluating}, recent systems like AlphaCode~\citep{li2022alphacode}, AlphaTensor~\citep{Fawzi2022}, and AlphaProof~\citep{alphaproof} have pushed the boundaries of reasoning and are able to discover novel solutions for fundamental problems. However, these systems typically rely on massive, closed-source models and significant computational resources. This dependency has sparked interest in orchestrating smaller, open-weight models~\citep{belcak2025smalllanguagemodelsfuture} to democratize access.

To surpass the limitations of standard prompting, recent research has successfully coupled LLMs with evolutionary algorithms, using models as ``mutation operators'' that iteratively refine a population of candidate programs. Pioneers like FunSearch~\citep{funsearch} and Evolution of Heuristics (EoH)~\citep{eoh} proved this synergy could generate new solutions that outperform human baselines. Building on this foundation, Google DeepMind released AlphaEvolve~\citep{alphaevolve_whitepaper}, capable of evolving entire codebases and producing state-of-the-art (SOTA) solutions to a diverse range of problems, yet its closed-source nature and reliance on proprietary models hinders reproducibility.

In this work, we introduce \method{}, an evolutionary coding agent that operationalizes LLM-driven search within a transparent, fully open framework. It tackles a meta-optimization problem in which candidate programs are iteratively refined through selection, variation, and recombination guided by execution feedback and fitness signals. Concretely, \method{} combines (i) an islands-based genetic algorithm for diversity and parallel search, (ii) a weighted LLM ensemble for model selection, and (iii) three modular operators for exploration and exploitation: inspiration-based crossover, meta-prompting, and depth-based targeted refinement. The key contribution does not lie in the individual components, but rather in the orchestration of these mechanisms into a reproducible search system that balances global exploration, local refinement, and executable feedback.

We evaluate \method{} on problems from the AlphaEvolve benchmark~\citep{alphaevolve_whitepaper} and the EoH benchmark suite~\citep{eoh}. Comparisons with AlphaEvolve and ThetaEvolve use their reported results, while comparisons with OpenEvolve and ShinkaEvolve are run under matched configurations and budgets. \method{} matches or surpasses reported AlphaEvolve results on 5 of 9 benchmarks and outperforms OpenEvolve and ShinkaEvolve in 6 of 9 problems under matched conditions. On the EoH suite, \method{} achieves competitive performance against EoH on heuristic design tasks using only default configurations and a single open-weight LLM. Ablations show that the synergy between \method{}'s components, rather than any single operator, is the key enabling factor for these results.

Our contributions include:
\begin{itemize}[noitemsep]
    \item An open-source framework for algorithmic discovery that integrates islands-based evolutionary search with modular LLM orchestration, designed for transparency and reproducibility.
    \item A comprehensive empirical evaluation on established algorithm-discovery benchmarks, including reported closed-source baselines and controlled open-source comparisons under matched budgets, demonstrating strong quality--cost trade-offs with open-weight models.
    \item An extensive ablation and component-level analysis showing that the synergy between \method{}'s operators is central to its strongest results, and that no single component suffices in isolation.
\end{itemize}

\section{Related Work}
\paragraph{Evolutionary Program Synthesis.}
Automated program optimization has its roots in Genetic Programming (GP), where populations of code are iteratively refined via crossover and mutation~\citep{Koza1994, langdon2013foundations}. Recently, LLMs have redefined this field by acting as semantically-aware operators, capable of generating high-quality solutions to complex tasks~\citep{li2022alphacode, alphaproof}. This paradigm, often termed ``Evolution through Large Models''~\citep{Lehman2024, Hemberg2024}, replaces traditional stochastic bit-flipping with informed, language-based reasoning.

\paragraph{Automatic heuristic design (AHD).}
FunSearch~\citep{funsearch} first demonstrated that coupling LLMs with evolutionary search can discover solutions surpassing human baselines. EoH~\citep{eoh} generalized this into a framework for evolving scoring functions or update rules that operate inside fixed algorithmic pipelines and has since become a widely adopted open-source AHD baseline, inspiring many subsequent methods~\cite{ReEvo,multiobj,metaheuri,liu2025eohsevolutionheuristicset}. AHD frameworks evolve only a single component within a fixed algorithm rather than synthesizing standalone programs end-to-end.

\paragraph{End-to-end algorithmic discovery.}
Google DeepMind's AlphaEvolve~\citep{alphaevolve_whitepaper, georgiev2025mathematicalexplorationdiscoveryscale} bridges this gap by synthesizing full programs in large-scale scenarios---including GPU kernel optimization and warehouse computing---but remains closed-source. In response, open-source projects such as OpenEvolve~\citep{openevolve}, ShinkaEvolve~\citep{lange2025shinka}, and ThetaEvolve~\citep{wang2025thetaevolvetesttimelearningopen} have emerged to democratize end-to-end algorithmic discovery, with domain-specific variants targeting scaling laws~\citep{lin2025languagemodelsdiscoverscaling}, cloud scheduling~\citep{cheng2025barbariansgateaiupending}, and hardware-specific optimizations~\citep{nagaitsev2025optimizingpytorchinferencellmbased}.

\paragraph{Meta-Optimization.}
A significant challenge in LLM-driven search is the sensitivity of models to minor prompt variations~\citep{anagnostidis2024susceptiblellmsinfluenceprompts, leviathan2025promptrepetitionimprovesnonreasoning}. To mitigate this, researchers have explored meta-prompting---using the LLM to manage its own reasoning instructions~\citep{suzgun2024metapromptingenhancinglanguagemodels, zhang2025metapromptingaisystems}---and evolutionary search for prompts, in which natural language strategies for certain problems are iteratively improved to increase the likelihood of an LLM to produce high-performing solutions~\citep{chen2023evoprompting, guo2025evopromptconnectingllmsevolutionary,agrawal2026gepareflectivepromptevolution}.

\method{} sits at the intersection of these trends, applying evolutionary strategies not just to solution programs, but to the search process itself. Unlike heuristic-design systems such as FunSearch and EoH, \method{} targets end-to-end program discovery, where candidate solutions are complete programs evaluated directly on the task rather than components inserted into a fixed solver. By mirroring the optimization of programs with the evolution of prompts, \method{} enables models to reflect on and rewrite their own instructions. This design yields more diverse search trajectories while prioritizing reproducibility and transparent evaluation.

\section{Preliminaries}
This section introduces the notation and abstractions used to describe the evolutionary process over solutions and prompts.

A \textbf{solution} is a program generated to solve a problem. When an existing solution $S$ is used in a prompt to generate a new solution $S'$, we say that $S$ is the \textit{parent solution} of $S'$. This parent-children relationship imposes a natural forest structure on the solution population, which is a collection of rooted, directed trees. For any solution $S$, we denote the set of its $k$ nearest ancestors as $A_k(S)$.

A \textbf{prompt} is a textual input provided to an LLM to generate a solution. We define the prompt used for generating solution 
$S$ as its \textit{parent prompt}, denoted by $P(S)$. Since LLMs are inherently probabilistic, a single prompt can generate multiple distinct solutions.

The quality of a solution is quantified by an \textbf{evaluation function,} $h:\mathcal{S} \mapsto \mathbb{R}^d$, which maps a solution $S$ from the space of all possible solutions $\mathcal{S}$ to a real-valued vector of performance metrics, such as runtime, memory usage, or objective value. 

We also define two \textbf{fitness functions} to measure the overall quality of prompts and solutions. The \textbf{solution fitness}, $f_{\mathrm{sol}}:\mathcal{S}\mapsto\mathbb{R}_{\geq 0}$, maps a solution to a non-negative score and typically corresponds to the primary metric in $h$ that we aim to maximize. The \textbf{prompt fitness}, $f_{\mathrm{prompt}}$, is derived from $f_{\mathrm{sol}}$ and is defined as the maximum fitness achieved by any solution generated from that prompt:
\begin{equation}\label{eq:f_prompt_def}
    f_{\mathrm{prompt}}(P) = \max_{S:P(S) = P}\{f_{\mathrm{sol}}(S)\}.
\end{equation}
This rewards prompts that have demonstrated the potential to generate high-quality solutions, making them valuable candidates for future evolution, even if some of their offspring may be suboptimal.

The primary \textbf{optimization goal} is to iteratively evolve an initial population of prompts and solutions, in order to maximize the solution fitness $f_{\mathrm{sol}}$ over a maximum number of epochs $N$, while respecting constraints on other metrics from $h$, such as execution time and memory usage.

\section{Methodology}\label{sec:methodology}
\begin{figure*}[!t]
  \includegraphics[width=1\linewidth]{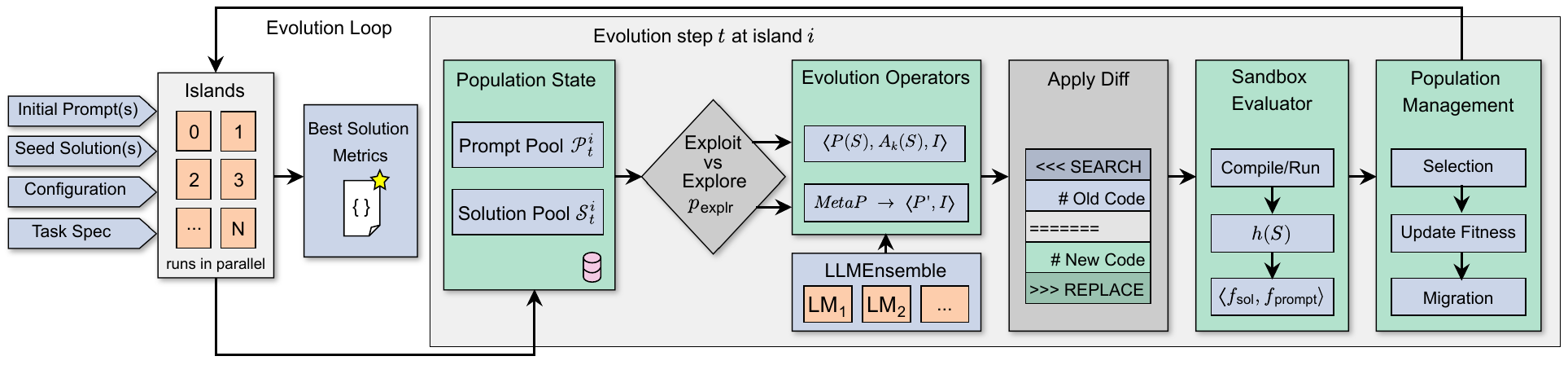}
  \caption{Overview of \method{}.}
\end{figure*}

\method{} integrates an evolutionary framework with LLMs to optimize programs. The architecture is based on the island genetic algorithm~\citep{islands_paper}: multiple populations (islands) evolve independently and periodically exchange their best-performing individuals (migration) according to a predefined topology. This design improves concurrent evaluation, maintains diversity, and propagates successful solutions across the search. At each epoch $t$, every island $i$ maintains a population of prompts $\mathcal{P}_t^i$ and solutions $\mathcal{S}_t^i$. \method{} operates through an iterative process that progressively enhances populations of prompts and solutions via three components. Evolutionary Operators (Section~\ref{sec:operators}) generate new individuals, balancing exploration and exploitation. An LLM Ensemble (Section~\ref{sec:ensemble}) provides the generative engine for code modification and recombination. Population Management (Section~\ref{sec:population}) handles evaluation, fitness tracking, migration, and archive updates.

\subsection{LLM Ensemble for Solution Generation}\label{sec:ensemble}
The engine behind \method{}'s solution generation is a weighted ensemble of LLMs---denoted \texttt{LLMEnsemble}---that modify and combine preexisting solutions. For each generation task, a model is sampled according to ensemble weights. The sampled model is then prompted to generate new solutions via edits using a diff-based \texttt{SEARCH/REPLACE} format: the model identifies a code region and proposes a targeted replacement. Users can configure distinct ensembles for exploration and exploitation (e.g., higher-temperature models for exploration; lower-temperature models for exploitation). In the simplest case, the ensemble consists of a single LLM.

\subsection{Evolutionary Operators}\label{sec:operators}
New solutions are generated in parallel using exploitation or exploration operators, sampled independently across islands. At each step, one operator is chosen according to an exploration rate $p_{\text{explr}}$, which is governed by a scheduler (Section~\ref{sec:scheduler}):
\begin{enumerate}[noitemsep,leftmargin=12pt]
    \item \textbf{Depth exploitation.} This operator refines high-performing solutions. A parent $S$ is selected from $\mathcal{S}_t^i$ via rank-based selection, with probability inversely proportional to its rank:
    \begin{equation}\label{eq:rank_sampling}
        \Pr(S) := \frac{\text{rk}(S)^{-1}}{\sum_{S' \in \mathcal{S}_t^i}\text{rk}(S')^{-1}},
    \end{equation}
    where $\text{rk}(S)$ is the position of $S$ when sorting $\mathcal{S}_t^i$ by $f_{\text{sol}}$ in descending order. The ensemble is prompted with $S$, its parent prompt $P(S)$, and its $k$ nearest ancestors $A_k(S)$. This truncated ancestral context encourages targeted, incremental improvements as depth increases, rather than wholesale strategy changes.
    \item \textbf{Meta-prompting exploration.} This operator fosters solution diversity and enriches the prompt population with feedback from previous solutions. A solution $S$ and a prompt $P$ are sampled independently and uniformly at random. An auxiliary LLM \texttt{MetaPromptingLLM} generates an enriched prompt $P'$ by analyzing $P$ and $S$. The \texttt{LLMEnsemble} uses $P'$ and $S$ to generate a new solution $S'$. We intentionally exclude the ancestor chain to allow exploration of novel strategies unconstrained by lineage, while leveraging the richer prompt $P'$.
\end{enumerate}

\paragraph{Inspiration-based Crossover.}
For both exploitation and exploration, we provide the ensemble with a set of ``inspiration'' solutions, which are sampled by rank in the former case (Eq.~\ref{eq:rank_sampling}) or uniformly at random in the latter case. Because \method{} adopts a diff-based \texttt{SEARCH/REPLACE} editing format---where the LLM identifies a code region in a single parent program and proposes a targeted replacement---traditional binary crossover is ill-suited: it would require identifying semantically compatible regions across two distinct programs and merging them into a coherent diff, which is highly error-prone. Instead, our inspiration-based approach achieves the semantic intent of crossover more naturally: the LLM is given multiple inspiration solutions alongside the parent, understands the high-level strategies they embody, and incorporates those ideas into the parent via targeted edits. This makes the operator effectively multi-parent, with binary crossover as the special case of a single inspiration.

Algorithm~\ref{algo:codeevolve} presents the core operator loop. The \texttt{LLMEnsemble} receives a prompt, an ancestor set (possibly empty), the target solution, and inspirations, and outputs a new solution. The \texttt{MetaPromptingLLM} receives a prompt and a solution and outputs a new prompt. For readability, the pseudocode omits the scheduler and population management logic, as both are orthogonal to the operator logic and described below.

\begin{algorithm}[tb]
\caption{Core exploitation/exploration loop of \method{} at epoch $t$}
\begin{algorithmic}[1]
\STATE \textbf{Input:} Populations $\mathcal{P}_t^i,\mathcal{S}_t^i$, exploration probability $p_{\text{explr}}$, maximum ancestor depth $k$
\STATE \textbf{Output:} New solution $S'$, and new prompt $P'$ if exploration is chosen
\STATE Sample $p \sim \textrm{Uniform}(0,1)$
\IF{$p < 1-p_{\text{explr}}$}
    \STATE Sample $S \in \mathcal{S}_t^i$ and inspirations $I \subseteq \mathcal{S}^i_t\setminus\{S\}$ according to Eq.~\ref{eq:rank_sampling}
    \STATE Collect ancestor solutions $A_k(S)$ from $S$ to its root in $\mathcal{S}^i_t$
    \STATE $S'\gets \texttt{LLMEnsemble}(P(S),A_k(S),S,I)$
    \STATE $P' \gets \NULL$
\ELSE 
    \STATE Sample $S \in \mathcal{S}_t^i$, inspirations $I \subseteq \mathcal{S}^i_t\setminus\{S\}$, and $P \in \mathcal{P}_t^i$ uniformly at random
    \STATE $P' \gets \texttt{MetaPromptingLLM}(P,S)$
    \STATE $S' \gets \texttt{LLMEnsemble}(P',\emptyset,S,I)$
\ENDIF
\RETURN $S',P'$
\end{algorithmic}
\label{algo:codeevolve}
\end{algorithm}

\subsection{Exploration Scheduling}\label{sec:scheduler}
\method{} controls the exploration--exploitation trade-off via a scheduler that adjusts the exploration probability $p_{\text{explr}}$ over time. In our experiments, we primarily use a \emph{Plateau Scheduler}, which monitors improvements in best-so-far fitness over a sliding window. When progress stalls, the scheduler temporarily increases $p_{\text{explr}}$ to encourage exploration; once improvement resumes, it gradually anneals back to a baseline value. We also implement standard decay-based schedules (e.g., exponential or cosine). The scheduler is lightweight, requiring only recent fitness statistics, and operates independently of the evolutionary operators themselves.

\subsection{Population Management}\label{sec:population}
\method{} includes three mechanisms to manage populations over time.

\textbf{Initialization.} The algorithm begins with an initial solution (often trivial, e.g., a function returning zero) and a basic prompt describing the problem. To create a diverse starting population at each island, the \texttt{LLMEnsemble} is prompted multiple times with this initial pair, generating independent approaches that become roots of new solution trees.

\textbf{Evaluation and Population Control.} Each new solution is executed in a sandbox with runtime and memory limits. If execution succeeds, we compute $f_{\text{sol}}(S)$ and metrics $h(S)$ and add $S$ to the population. Failures receive fitness zero, with logs stored for instructive context in future prompts. 

\textbf{Elitist Migration.} Top performers from each island are copied to neighboring islands at a fixed migration frequency and rate. To prevent cycles and premature convergence, a solution migrates at most once from its origin island, and we never migrate the best-performing solution of an island in order to preserve its uniqueness. Migrants become roots of new trees upon arrival, with parent pointers set to \texttt{NULL}.

\paragraph{Quality--diversity via MAP-Elites.}
In addition to island populations, \method{} maintains a per-island MAP-Elites archive~\citep{mouret2015illuminatingsearchspacesmapping}. Users define feature descriptors (e.g., code-level properties, algorithmic behaviors, or runtime profiles). The archive partitions the feature space either as a regular lattice (MAP-Elites) or via centroidal Voronoi tessellations (CVT-MAP-Elites)~\citep{vassiliades2017usingcentroidalvoronoitessellations}, storing the elite (i.e., most fit) solution per cell. Archive updates occur after evaluation: each successful solution $S$ is mapped to its feature cell, and if $S$ improves the cell’s fitness, it replaces the incumbent elite. Sampling for inspirations or parents then draws from the archive to inject structured diversity, e.g., proportional to the rank of all elites in the grid or uniformly over filled cells. The archive thus complements island dynamics by probing diverse niches and reducing premature convergence.

Together, these components are mutually reinforcing: depth exploitation narrows the search along proven lineages, providing stable scaffolding for incremental refinement; inspiration-based crossover prevents premature convergence by importing strategies from distant archive cells that depth exploitation would otherwise never reach; and CVT-MAP-Elites ensures the diversity pool remains populated throughout, so that crossover always has meaningful genetic material to draw from. Our ablations (Section~\ref{sec:experiments}) confirm that no single component suffices in isolation---it is their interaction that yields the strongest performance.

\section{Experiments}\label{sec:experiments}
In this section, we are interested in answering the following questions:
\begin{enumerate}[noitemsep,label=\textbf{RQ\arabic*}]
    \item Can \method{} advance the state-of-the-art in automated algorithmic discovery?
    \item Can smaller open-weight models compete with more expensive, closed-source LLMs as the backbone of \method{}?
    \item How do the individual components of \method{} contribute to performance and sample efficiency on the proposed benchmarks?
\end{enumerate}

Our primary evaluation is built around a selection of problems from the AlphaEvolve benchmark~\citep{alphaevolve_whitepaper}, which defines challenging baselines for automated algorithmic discovery. Apart from comparing \method{} against AlphaEvolve itself, which represents the state-of-the-art based on proprietary frontier LLMs, we also compare it against ThetaEvolve~\citep{wang2025thetaevolvetesttimelearningopen}, an RL-based framework that uses open-weight models and is able to produce SOTA solutions to some of the problems at hand. Because AlphaEvolve is closed-source and ThetaEvolve requires large-scale RL infrastructure, these comparisons rely on reported numbers. We complement this with controlled comparisons against OpenEvolve~\citep{openevolve} and ShinkaEvolve~\citep{lange2025shinka}---the two most prominent open-source frameworks for general algorithmic discovery---under matched configurations and evaluation budgets, assessing \method{}'s performance in a fair head-to-head setting. We additionally evaluate on the EoH benchmark suite~\citep{eoh} to demonstrate breadth on heuristic design tasks.

\subsection{Benchmark Problems}
We evaluate \method{} on problems drawn from two suites. From the AlphaEvolve benchmark~\citep{alphaevolve_whitepaper} we use three packing problems (\circlepacking{}, \texttt{CirclePackingRect}, \texttt{HexagonPacking}), a point-placement problem (\minmax{}), and two autocorrelation-inequality problems (\texttt{FirstAutocorrIneq}, \texttt{SecondAutocorrIneq}), yielding nine instances in total. From the Evolution of Heuristics (EoH) benchmark suite~\citep{eoh} we use the Online Bin Packing Problem (OBPP), the Traveling Salesman Problem (TSP), and the Flow Shop Scheduling Problem (FSSP). Formal definitions, parameter settings, and pointers to the original implementations are deferred to Appendix~\ref{appendix:benchmarks}.

\subsection{Experimental setup}
All experiments were conducted on \textit{AWS SageMaker} with a fixed per-run computational budget in terms of vCPUs and RAM, with each candidate solution being evaluated in an isolated sandbox with strict runtime and memory constraints (see Appendix~\ref{appendix:configs} for more details). We evaluate two ensemble configurations: one using Google's \texttt{GEMINI-2.5} models~\cite{gemini} to approximate a frontier-model setting, and another using only Qwen's \texttt{Qwen3-Coder-30B}~\citep{yang2025qwen3technicalreport} to explore the performance–cost trade-off with open-weight models. Experiments with other LLM backbones can be found in Appendix~\ref{appendix:other_llms}, and representative prompts and evolved solutions are shown in Appendix~\ref{appendix:prompts_sols}.

\subsection{Main Results}
\begin{table*}[tb]
    \centering
    \small
    \caption{Comparison between best solutions of \method{} and reported best solutions of AlphaEvolve~\citep{alphaevolve_whitepaper} and ThetaEvolve~\citep{wang2025thetaevolvetesttimelearningopen}. Best results are in \textbf{bold}.}
    \label{tab:framework_results}
    % \resizebox{\textwidth}{!}{
    \begin{tabular}{l|c|c|cc}
        \toprule
        \multirow{2}{*}{\textbf{Problem}}& \multirow{2}{*}{\textbf{AlphaEvolve}} & \textbf{ThetaEvolve} &\multicolumn{2}{c}{\textbf{\method{}}}  \\ 
        &  & {\scriptsize \texttt{Distill-Qwen3-8B}}  &  {\scriptsize \texttt{Qwen3-Coder-30B}} & {\scriptsize \texttt{GEMINI-2.5 FLASH/PRO}} \\ 
        
        \midrule
        
        \texttt{CirclePackingSquare($n=26$)} ($\uparrow$)       & $2.63586$ & $\mathbf{2.63598}$  & $\mathbf{2.63598}$ & $2.63597$ \\
        \texttt{CirclePackingSquare($n=32$)} ($\uparrow$)     & $2.93794$  & --- & $\mathbf{2.93956}$ & $2.93950$ \\
        \texttt{CirclePackingRect($n=21$)} ($\uparrow$)         & $\mathbf{2.36583}$ & ---  & $2.36339$ & $\mathbf{2.36583}$ \\
        \texttt{HexagonPacking($n=11$)} ($\downarrow$)             & $\mathbf{3.93009}$ & --- & $4.02786$ & $3.93794$ \\
        \texttt{HexagonPacking($n=12$)} ($\downarrow$)             & $\mathbf{3.94191}$ & --- & $4.01092$ & $4.00001$ \\
        \hline
        \texttt{MinimizeMaxMinDist($n=16, d=2$)} ($\downarrow$)    & $12.88927$ & --- & $12.88925$ & $\mathbf{12.88923}$ \\
        \texttt{MinimizeMaxMinDist($n=14, d=3$)} ($\downarrow$)   & $4.16585$ & ---  & $4.16765$ & $\mathbf{4.16579}$ \\
        \hline
        \texttt{FirstAutocorrIneq} ($\downarrow$)                   & $1.50316$ & $\mathbf{1.50313}$ & $1.51343$ & $1.55438$ \\
        \texttt{SecondAutocorrIneq} ($\uparrow$)              & $\mathbf{0.96102}$ & $0.94690$ & $0.88110$ & $0.87067$  \\
        
        \bottomrule
    \end{tabular}
    % }
\end{table*}

\begin{table*}[tb]
\centering
\small
\caption{Controlled comparison of \method{}, ShinkaEvolve~\citep{lange2025shinka}, and OpenEvolve~\citep{openevolve} using \texttt{Qwen3-Coder-30B} across three independent runs with matched configurations and evaluation budgets.}
\label{tab:comp_shinka_open}
\begin{tabular}{l|cc|cc|cc}
\toprule
\textbf{Problem} & \multicolumn{2}{c|}{\textbf{ShinkaEvolve}} & \multicolumn{2}{c|}{\textbf{OpenEvolve}} & \multicolumn{2}{c}{\textbf{\method{}}} \\
& \textbf{Mean} & \textbf{Best} & \textbf{Mean} & \textbf{Best} & \textbf{Mean} & \textbf{Best} \\
\midrule
\texttt{CirclePackingSquare($n=26$)}($\uparrow$) & 2.47568 & 2.61551 & 2.54962 & 2.6245 & \textbf{2.63576} & \textbf{2.63598} \\
\texttt{CirclePackingSquare($n=32$)}($\uparrow$) & 2.53205 & 2.69473 & 2.84897 & 2.93156 & \textbf{2.93751} & \textbf{2.93957} \\
\texttt{CirclePackingRect($n=21$)}($\uparrow$)  & 2.21694 & 2.32701 & 2.28131 & \textbf{2.36513} & \textbf{2.36212} & 2.36465 \\
\hline
\texttt{HexagonPacking($n=11$)}($\downarrow$)  & 5.55234 & 5.00000 & 4.30696 & 4.20679 & \textbf{4.02491} & \textbf{4.01377} \\
\texttt{HexagonPacking($n=12$)}($\downarrow$) & 5.07507 & 4.38608 & \textbf{4.01991} & \textbf{4.0001} & 4.10841 & 4.05198 \\
\hline
\texttt{MinimizeMaxMinDist($n=16, d=2$)}($\downarrow$) & 13.48234 & 13.11627 & 12.97046 & 12.94042 & \textbf{12.91569} & \textbf{12.90274} \\
\texttt{MinimizeMaxMinDist($n=14, d=3$)}($\downarrow$)  & 4.37160 & 4.21087 & \textbf{4.18983} & 4.1714 & 4.26257 & \textbf{4.16765} \\
\hline
\texttt{FirstAutocorrIneq}($\downarrow$) & \textbf{1.52173} & \textbf{1.51556} & 1.58031 & 1.5751 & 1.54696 & 1.51642 \\
\texttt{SecondAutocorrIneq}($\uparrow$) & 0.82010 & 0.84646 & 0.84634 & 0.86086 & \textbf{0.86533} & \textbf{0.90114} \\
\bottomrule
\end{tabular}
\end{table*}

\paragraph{Advancing the state-of-the-art.} Table~\ref{tab:framework_results} compares the best results of \method{} against the reported best results of AlphaEvolve and ThetaEvolve. \method{} matches or surpasses AlphaEvolve on 5 of the 9 benchmark instances, and establishes new best-known results on \minmax{} and \circlepacking{} ($n=32$). The main gap is on the autocorrelation inequalities, where both AlphaEvolve and ThetaEvolve achieve substantially better results. Appendix~\ref{appendix:sol_viz} shows visualizations of representative solutions found by \method{} and AlphaEvolve (Figures~\ref{fig:results_circlepackingsquare}-\ref{fig:results_minmaxmindist3}).

\paragraph{Controlled comparison with open-source frameworks.} Table~\ref{tab:comp_shinka_open} evaluates \method{} against OpenEvolve and ShinkaEvolve under identical configurations, evaluation budgets, and LLM backbone (see Appendix~\ref{appendix:other_os_frameworks} for full details). \method{} achieves the best solution in 6 of 9 problems, and the best mean fitness also in 6 of 9 problems, showing strong performance among open-source frameworks under matched conditions.

\paragraph{Comparison with EoH on heuristic design.}
EoH evolves a single callable component within a fixed algorithmic pipeline using fixed prompt templates---a design optimized for automatic heuristic design (AHD) rather than end-to-end program discovery. Since EoH's architecture makes a direct configuration match infeasible, we compare against EoH's reported results---obtained with GPT-3.5-turbo---while matching the total number of LLM calls and evaluation budget using \texttt{Qwen3-Coder-30B}. We ran \method{} with $10$ islands: 154 epochs for OBPP (30s timeout), 152 for TSP (60s), and 152 for FSSP (120s), reporting best and mean over three independent runs. Complete configurations are available at \codeevolvemainrepo.

\begin{table*}[tb]
\centering
\small
\caption{Comparison between \method{} and EoH's reported results~\citep{eoh} for the Online Bin Packing Problem (OBPP). Lower is better.}
\label{tab:eoh_bin_packing}
\begin{tabular}{l|c|c|c|c|c|c}
\toprule
Method & 1k C100 & 5k C100 & 10k C100 & 1k C500 & 5k C500 & 10k C500 \\
\midrule
EoH & \textbf{2.24\%} & \textbf{0.80\%} & \textbf{0.61\%} & 2.13\% & 0.78\% & 0.61\% \\
\method{} (Best) & 4.18\% & 3.16\% & 3.22\% & \textbf{0.00\%} & \textbf{0.20\%} & \textbf{0.20\%} \\
\method{} (Mean) & 4.34\% & 3.34\% & 3.38\% & \textbf{0.17\%} & \textbf{0.37\%} & \textbf{0.36\%} \\
\bottomrule
\end{tabular}
\end{table*}
\begin{table}[tb]
\centering
\small
\caption{Comparison between \method{} and EoH's reported results~\citep{eoh} for the Traveling Salesman Problem (TSP). Lower is better.}
\label{tab:eoh_tsp}
\begin{tabular}{l|c|c|c}
\toprule
Method & TSP20 & TSP50 & TSP100 \\
\midrule
EoH & \textbf{0.000} & \textbf{0.000} & \textbf{0.025} \\
\method{} (Best) & \textbf{0.000} & 0.001 & 0.040 \\
\method{} (Mean) & \textbf{0.000} & 0.011 & 0.101 \\
\bottomrule
\end{tabular}
\end{table}
\begin{table}[tb]
\centering
\small
\caption{Comparison between \method{} and EoH's reported results~\citep{eoh} for the Flow Shop Scheduling Problem (FSSP). Lower is better.}
\label{tab:eoh_fssp}
\begin{tabular}{l|c|cc}
\toprule
\multirow{2}{*}{\textbf{Instance}} & \multirow{2}{*}{\textbf{EoH}} & \multicolumn{2}{c}{\textbf{\method{}}} \\
 & & \textbf{Best} & \textbf{Mean} \\
\midrule
n20m5 & 0.09 & \textbf{0.02} & 0.33 \\
n20m10 & 0.30 & \textbf{0.19} & 0.25 \\
n20m20 & \textbf{0.10} & 0.11 & 0.14 \\
n50m5 & \textbf{0.02} & 0.04 & 0.05 \\
n50m10 & \textbf{0.19} & 0.27 & 0.28 \\
n50m20 & \textbf{0.60} & 0.69 & 0.72 \\
n100m5 & -0.04 & \textbf{-0.05} & -0.03 \\
n100m10 & 0.14 & \textbf{0.12} & 0.15 \\
n100m20 & 0.41 & \textbf{0.40} & 0.47 \\
n200m10 & \textbf{0.12} & 0.19 & 0.23 \\
n200m20 & 0.61 & \textbf{0.57} & 0.65 \\
\midrule
Average & \textbf{0.23} & \textbf{0.23} & 0.29 \\
\bottomrule
\end{tabular}
\end{table}

Tables~\ref{tab:eoh_bin_packing},~\ref{tab:eoh_tsp}, and~\ref{tab:eoh_fssp} show the comparison. On OBPP, \method{} significantly outperforms EoH on the out-of-distribution C500 instances, while EoH leads on C100. On TSP, \method{} matches EoH on TSP20 and is slightly behind on TSP50 and TSP100. On FSSP, \method{}'s best run matches EoH's overall average (0.23 gap) and beats EoH on 6 of 11 Taillard test sets. Taken together with the reported-best comparison in Table~\ref{tab:framework_results} and the controlled comparison in Table~\ref{tab:comp_shinka_open}, these results answer \textbf{RQ1} in the affirmative.

\paragraph{Cost efficiency.} The two ensemble configurations are complementary across problem types. \texttt{Qwen3-Coder-30B} achieves the strongest results on the \circlepacking{} instances and both autocorrelation problems, while \texttt{GEMINI-2.5} performs best on \texttt{CirclePackingRect}, both \texttt{HexagonPacking} instances, and both \minmax{} instances. The cost difference is most pronounced on \circlepacking{} ($n=26$): \texttt{Qwen3-Coder-30B} surpassed AlphaEvolve after roughly $900$ model calls at approximately $\$6$, whereas \texttt{GEMINI-2.5} required approximately $400$ calls at ${\sim}\$35$ (Figures~\ref{fig:sol_hist_circlepackingsquare26}-\ref{fig:sol_hist_circlepackingsquare32}). This order-of-magnitude cost advantage, combined with competitive performance overall, indicates that open-weight models under modular orchestration are often a viable and cost-effective alternative to frontier proprietary APIs, answering \textbf{RQ2} in the affirmative. A full cost and runtime breakdown is provided in Appendix~\ref{appendix:cost_and_runtime}.

\subsection{Ablations}
We ran ablation studies across all AlphaEvolve benchmark problems using \texttt{Qwen3-Coder-30B}. In every case, the full \method{} configuration produced the best peak performance, with complete plots provided in Appendix~\ref{appendix:further_ablations} (Figures~\ref{fig:depthinsp_ablation_circlepackingsquare}--\ref{fig:appendix_ablations_top}). In all such plots, the y-axis shows fitness on a $-\log(M-y+\epsilon)$ scale for increased visual clarity, where $M$ is the maximum fitness attained across all experiments, $y$ is the best fitness at each evaluation step, and $\epsilon$ is a small constant.

In the remainder of this section we focus on the \circlepacking{} problem with $n=32$, which serves as a standard reference benchmark in recent related work~\citep{openevolve,lange2025shinka,wang2025thetaevolvetesttimelearningopen} and offers an efficient testbed for isolating the contribution of individual components. We report the mean, best and worst outcomes across runs. 

\begin{figure*}[tb]
\centering
  \includegraphics[width=0.48\linewidth]{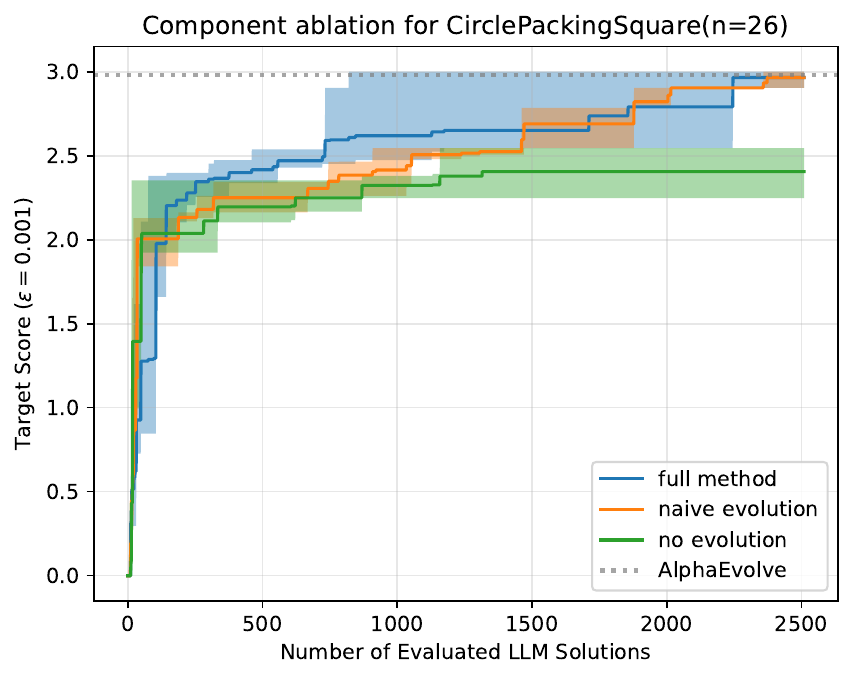} \hfill
  \includegraphics[width=0.48\linewidth]{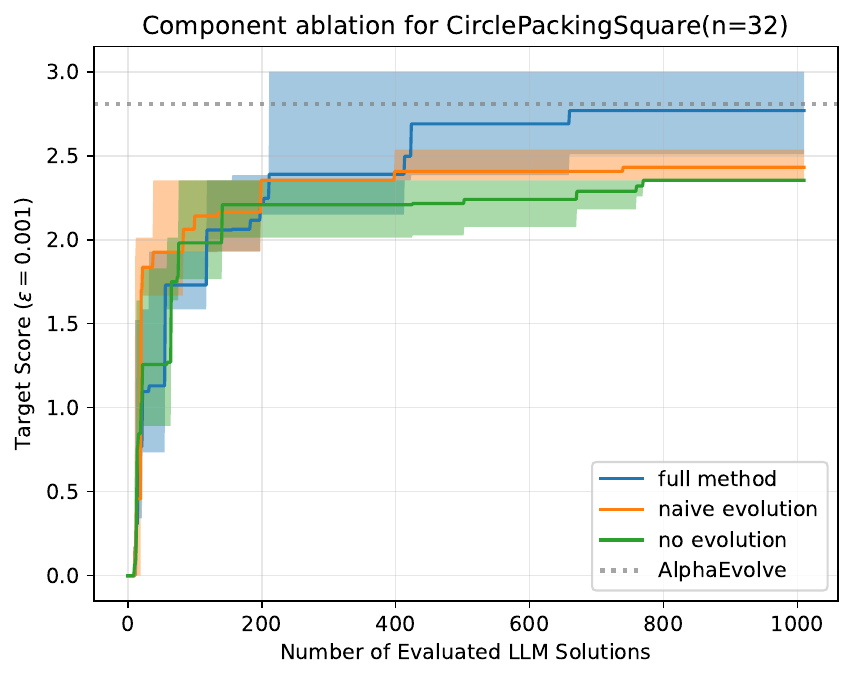}
  \caption {Component ablations for \method{} using \texttt{Qwen3-Coder-30B} on the \circlepacking{} problem. Curves show the mean across three distinct runs, and shaded regions show the best and worst results across all runs.}
  \label{fig:comp_ablation_circlepackingsquare}
\end{figure*}

\paragraph{Impact of components.}
We evaluated \method{} on three configurations: (i) ``full method'', which utilizes all operators described in Section~\ref{sec:operators}; (ii) ``naive evolution'', which uses a standard exploration/exploitation pipeline without the proposed components; and (iii) ``no evolution'', which repeatedly prompts the LLM with the initial prompt and solution, with no contextual data from other solutions. The full method outperforms both alternatives in mean performance and sample efficiency across all benchmarks (Figure~\ref{fig:comp_ablation_circlepackingsquare}). For \circlepacking{} ($n=32$), it is the only configuration that surpasses AlphaEvolve, while for $n=26$ the naive baseline requires more than twice as many evaluations to match the full method. This shows that, regarding \textbf{RQ3}, \method{}'s components not only increase overall performance, but are the enabling factor for obtaining state-of-the-art results.

\paragraph{Impact of depth and inspirations.}
We study two key operator hyperparameters: maximum ancestor depth $k$ and number of inspiration solutions $\iota$. Depth-only configurations ($\iota=0$) do not exceed AlphaEvolve, while inspiration-only ($k=0$, $\iota=2,3$) do. The full method consistently outperforms both in isolation, with higher sample efficiency, indicating a positive synergy between the two operators (Figure~\ref{fig:depthinsp_ablation_circlepackingsquare}). Further ablation studies (Appendix~\ref{appendix:further_ablations}) show that MAP-Elites and the choice of migration topology are also key factors for surpassing AlphaEvolve, reinforcing the positive effect of using \method{}'s components in unison. See Appendix~\ref{appendix:search_behavior} for additional analyses.

\paragraph{Practical guidelines and cross-backbone robustness.}
The same compact set of defaults proved robust across all benchmarks and the six LLM backbones we tested (Appendix~\ref{appendix:other_llms}): ring (\texttt{Cycle}) migration topology with migration rate $0.1$, initial exploration rate $0.2$ with the Plateau Scheduler, CVT-MAP-Elites with fitness and evaluation time as descriptors, maximum ancestor depth $k \in [3,5]$, and $\iota \in \{2,3\}$ inspirations. No prompt or operator retuning was needed when switching among \texttt{Qwen3-Coder-30B}, \texttt{GEMINI-2.5}, \texttt{GPT-4.1}, \texttt{Claude Sonnet 4.5}, \texttt{o4-mini}, and \texttt{gpt-oss-20B}, indicating that the orchestration benefits are not specific to a single model family. We provide a full practitioner-oriented discussion of budget-bound parameters, operator-level parameters, and robust defaults in Appendix~\ref{appendix:hyperparam_sensitivity}.

\section{Conclusion}
We introduced \method{}, an open-source framework that democratizes the search for novel algorithms. By integrating an islands-based genetic algorithm with modular LLM operators, \method{} bridges the gap between opaque, large-scale systems and accessible research tools. Our experiments demonstrate that \method{} matches or surpasses the reported AlphaEvolve results on multiple benchmark instances, compares favorably against open-source evolutionary frameworks under matched conditions, and achieves competitive results against EoH on heuristic design benchmarks despite not being specialized for automatic heuristic design. Perhaps most significantly, our results show that open-weight models, when properly orchestrated, offer a transparent and cost-effective alternative to proprietary APIs. 

% \method{} provides a foundation for future work in automated scientific discovery, enabling the community to iterate on search strategies and model ensembles within a reproducible framework.
% \clearpage

\clearpage\clearpage
\section*{Limitations}
While \method{} achieves best-known results on several benchmarks, important limitations remain.
First, comparisons with AlphaEvolve and ThetaEvolve rely on reported numbers because AlphaEvolve is closed-source and ThetaEvolve requires RL fine-tuning on infrastructure beyond our budget; differences in evaluation budgets, parallelism, or implementation details may affect these comparisons despite our efforts to match conditions.
Second, \method{} underperforms on the autocorrelation inequality benchmarks, where both AlphaEvolve and ThetaEvolve achieve substantially better results; this suggests that problems requiring highly specialized analytic constructions may be less amenable to \method{}'s combinatorial search approach and could benefit from domain-specific adaptations or stronger backbone models.
Third, a full ablation grid over problems, components, and LLM backbones is combinatorially expensive; we therefore concentrate the most detailed ablations on \texttt{Qwen3-Coder-30B}, while the cross-backbone results reported in Section~\ref{sec:experiments} and Appendix~\ref{appendix:other_llms} provide complementary evidence that the architectural benefits are not model-specific.
Fourth, the framework introduces hyperparameters (e.g., migration topology, number of islands, number of inspirations, maximum ancestor depth) that may require calibration on novel domains; we provide robust defaults (Appendix~\ref{appendix:hyperparam_sensitivity}), but cannot guarantee they transfer to all problem types.
Finally, while \method{} significantly reduces costs relative to closed-source alternatives, the inference budget for large-scale evolution remains non-trivial, potentially limiting accessibility for researchers with constrained compute resources.

% \section*{Acknowledgments}
% The authors thank Bruno Grossi for reviewing this
% paper and for the continuous support during the development of this project. We also thank Fernando Augusto and Tiago Machado for the useful conversations about possible applications of \method{} at Inter.

% \section*{Author Contributions (CRediT)}
% \textbf{Henrique Assumpção:} Conceptualization, Methodology, Software, Validation, Investigation, Data Curation, Writing – Original Draft, Writing – Review \& Editing, Project Administration. \textbf{Diego Ferreira:} Software, Validation, Investigation, Writing – Original Draft. \textbf{Leandro Campos:} Conceptualization, Formal Analysis, Writing – Review \& Editing. \textbf{Fabricio Murai:} Conceptualization, Formal Analysis, Writing – Original Draft, Writing – Review \& Editing.

\bibliography{bibliography}
\appendix
\section{Benchmark Problems}\label{appendix:benchmarks}
In this section, we provide further details about the problems used to evaluate \method{}. Section~\ref{appendix:benchmarks_ae} covers the four problem categories from the AlphaEvolve benchmark suite~\citep{alphaevolve_whitepaper} (nine instances in total), and Section~\ref{appendix:benchmarks_eoh} covers the three problems from the Evolution of Heuristics (EoH) benchmark suite~\citep{eoh}. The original EoH implementations are available in EoH's official repository,\footnote{\url{https://github.com/FeiLiu36/EoH}} and our adaptation of these problems for \method{} is available in \codeevolvemainrepo.

\subsection{AlphaEvolve benchmark}\label{appendix:benchmarks_ae}

\paragraph{Packing Circles and Hexagons.}
We consider three packing problems: \circlepacking{}, \texttt{CirclePackingRect}, and \texttt{HexagonPacking}. The first places $n$ disjoint unit circles in a unit square to maximize the sum of radii ($n=26,32$). The second places $n=21$ circles in a rectangle of perimeter $4$. The third packs $n=11,12$ unit regular hexagons into a larger hexagon while minimizing its side length.

\paragraph{Minimizing ratio of maximum to minimum distance.}
This benchmark consists of placing $n$ $d$-dimensional points in order to minimize the ratio between their maximum to minimum distance, referred to as \minmax{}, with instances $n=16,d=2$ and $n=14,d=3$.

\paragraph{First Autocorrelation Inequality.}
Let $C_1$ be the largest constant such that
\[
\max_{-1/2 \,\leq\, t \,\leq\, 1/2}(f*f)(t) \geq C_1 \left(\int_{-1/4}^{1/4}f(x)dx\right)^2,
\]
for all nonnegative functions $f:\mathbb{R} \mapsto \mathbb{R}$, where $f * f$ denotes the convolution operation. Upper bounds on $C_1$ can be obtained by explicitly constructing step-functions~\citep{matolcsi2009improvedboundssupremumautoconvolutions}, thus we wish to create an algorithm that generates a nonnegative step function that minimizes the ratio between $\max_{-1/2 \leq t \leq 1/2}(f*f)(t)$ and $(\int_{-1/4}^{1/4}f(x)dx)^2$.

\paragraph{Second Autocorrelation Inequality.}
Let $C_2$ be the smallest constant satisfying
\[
\|f*f\|_2^2 \leq C\|f*f\|_1\|f*f\|_\infty,
\]
for all nonnegative functions $f:\mathbb{R} \mapsto \mathbb{R}$, where $\|\cdot\|_p$ denotes the $p$-norm of a given function, and $f * f$ denotes the convolution operation. Hölder's inequality immediately yields $C_2 \leq 1$, and mathematicians have been attempting to bound $C_2$ from below by constructing explicit step functions~\citep{matolcsi2009improvedboundssupremumautoconvolutions}. The task at hand is thus to create an algorithm that generates a nonnegative step function that maximizes the ratio between $\|f*f\|_2^2$ and $\|f*f\|_1\|f*f\|_\infty$.

\subsection{EoH benchmark}\label{appendix:benchmarks_eoh}

\paragraph{Online Bin Packing Problem (OBPP).}
The OBPP problem~\citep{obpp} consists of allocating a collection of items of different sizes into the fewest possible bins with a given fixed capacity, where items are packed as they arrive, i.e., we do not have access to all items beforehand. Following~\citep{eoh}, the fitness is set to be the ratio between a theoretical optimal number of bins and the actual number of bins used, averaged over different instances.

\paragraph{Traveling Salesman Problem (TSP).}
The TSP problem~\citep{tsp} is among the most famous problems in computer science, and consists of finding a Hamiltonian cycle of minimum weight in a given weighted graph. Following~\citep{eoh}, the aim is to minimize the gap between the solution weight and the optimal solution generated by an exact TSP solver.

\paragraph{Flow Shop Scheduling Problem (FSSP).}
The FSSP problem~\citep{fssp} consists of allocating $n$ jobs on $m$ machines, where each job has $m$ tasks that must be performed in a predetermined order on the respective machines. Following~\citep{eoh}, the objective is to minimize the total schedule length.

\section{Supplemental experiments and results}\label{appendix:further_exp}

\subsection{Solution Visualizations}\label{appendix:sol_viz}
\begin{figure}[tb]
  \includegraphics[width=\columnwidth]{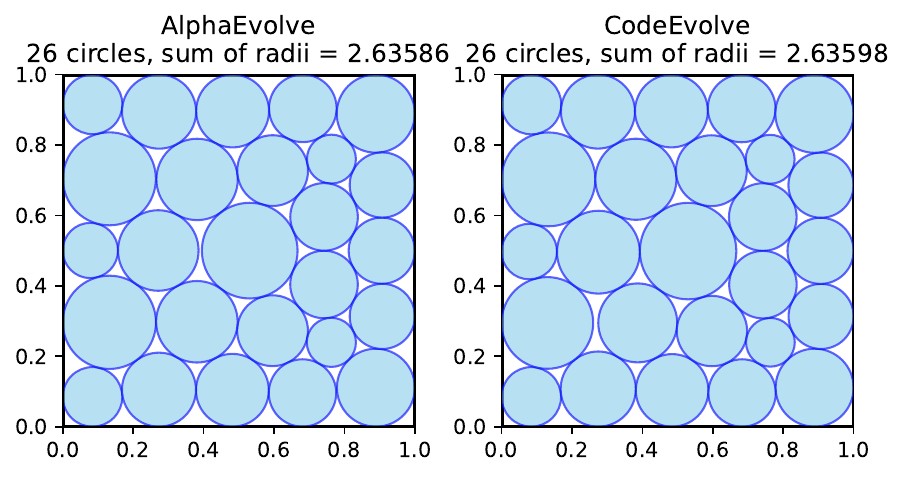}
  \caption{Comparison of best solutions found by \method{} and AlphaEvolve for the \circlepacking{} problem with $n = 26$.}
  \label{fig:results_circlepackingsquare}
\end{figure}

\begin{figure}[tb]
  \includegraphics[width=0.95\columnwidth]{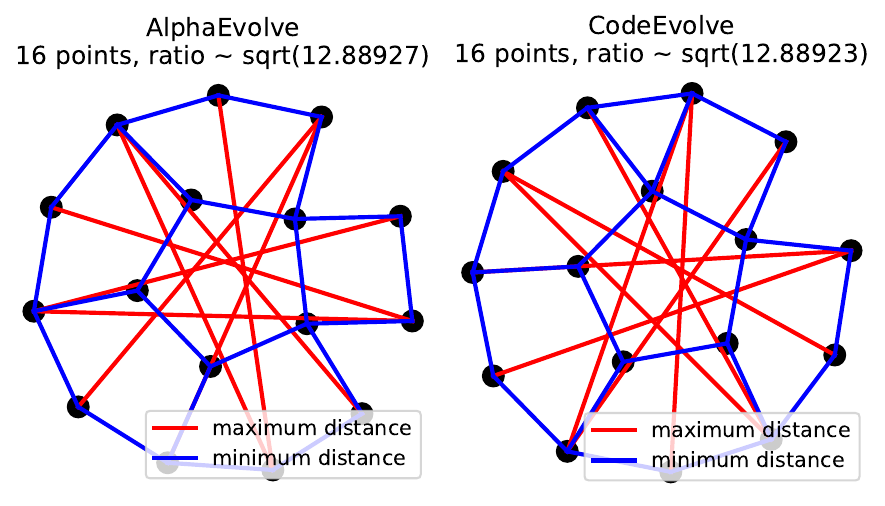}
  \caption {Comparison of best solutions found by \method{} and AlphaEvolve for the \minmax{} problem with $n=16,d=2$.}
  \label{fig:results_minmaxmindist}
\end{figure}

\begin{figure}[tb]
  \includegraphics[width=\columnwidth]{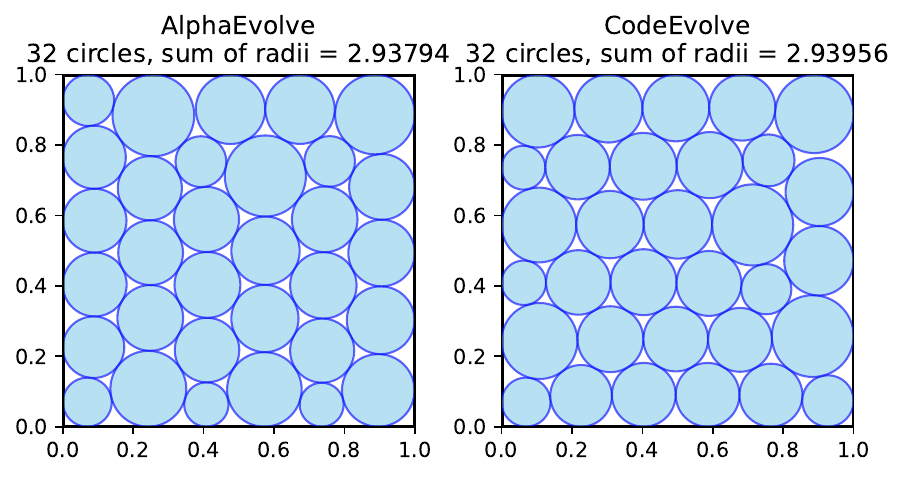}
  \caption{Comparison of best solutions found by \method{} and AlphaEvolve for the \circlepacking{} problem with $n = 32$.}
  \label{fig:results_circlepackingsquare32}
\end{figure}

\begin{figure}[tb]
  \includegraphics[width=\columnwidth]{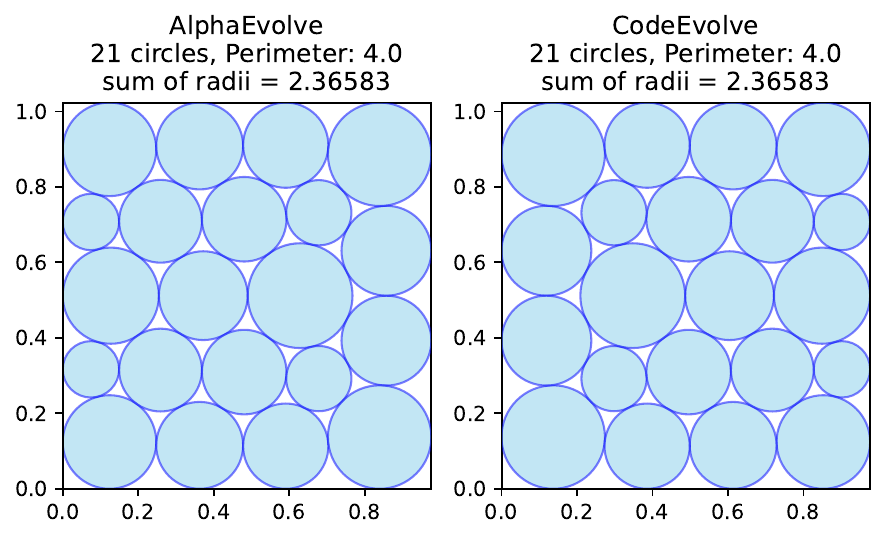}
  \caption{Comparison of best solutions found by \method{} and AlphaEvolve for the \texttt{CirclePackingRect} problem with $n = 21$.}
  \label{fig:results_circlepackingrect21}
\end{figure}

\begin{figure}[tb]
  \includegraphics[width=0.95\columnwidth]{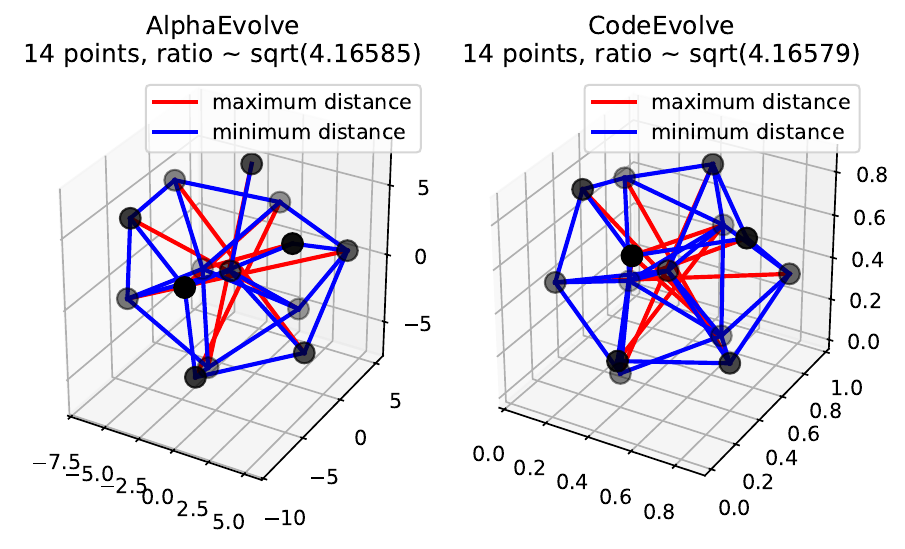}
  \caption {Comparison of best solutions found by \method{} and AlphaEvolve for the \minmax{} problem with $n=14,d=3$.}
  \label{fig:results_minmaxmindist3}
\end{figure}

Figures~\ref{fig:results_circlepackingsquare}--\ref{fig:results_minmaxmindist3} display representative best solutions found by \method{} and AlphaEvolve for packing and distance optimization benchmarks.

\subsection{Further ablations}\label{appendix:further_ablations}
\begin{figure*}[tb]
\centering
  \includegraphics[width=0.48\linewidth]{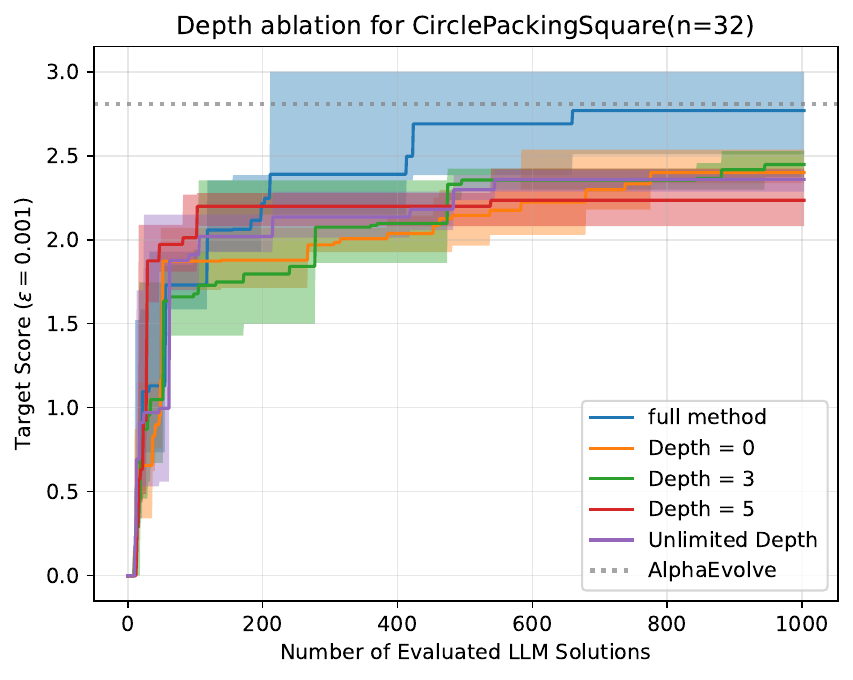} \hfill
  \includegraphics[width=0.48\linewidth]{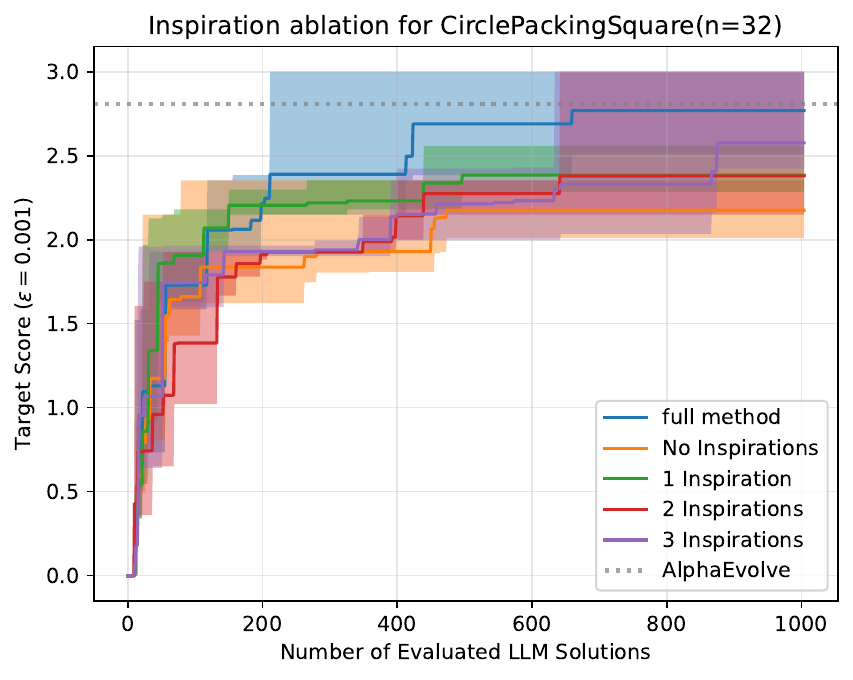}
  \caption {Depth and Inspiration ablations for \method{} using \texttt{Qwen3-Coder-30B} on the \circlepacking{} problem with $n = 32$.}
  \label{fig:depthinsp_ablation_circlepackingsquare}
\end{figure*}

\paragraph{Component ablations for \circlepacking{}.}
Figure~\ref{fig:comp_ablation_circlepackingsquare} shows the three-way component ablation (full method, naive evolution, no evolution) on \circlepacking{} for both $n=26$ and $n=32$. For $n=32$, only the full method surpasses the reported AlphaEvolve score, while for $n=26$ the naive baseline requires more than twice as many evaluations to match the full method. Figure~\ref{fig:depthinsp_ablation_circlepackingsquare} isolates the contributions of depth exploitation and inspiration-based crossover on the $n=32$ instance: depth-only does not exceed the reported AlphaEvolve score, inspiration-only ($\iota=2,3$) does, and combining both yields the highest sample efficiency---confirming the synergy discussed in Section~\ref{sec:experiments}.

\begin{figure*}[tb]
\centering
  \includegraphics[width=0.48\linewidth]{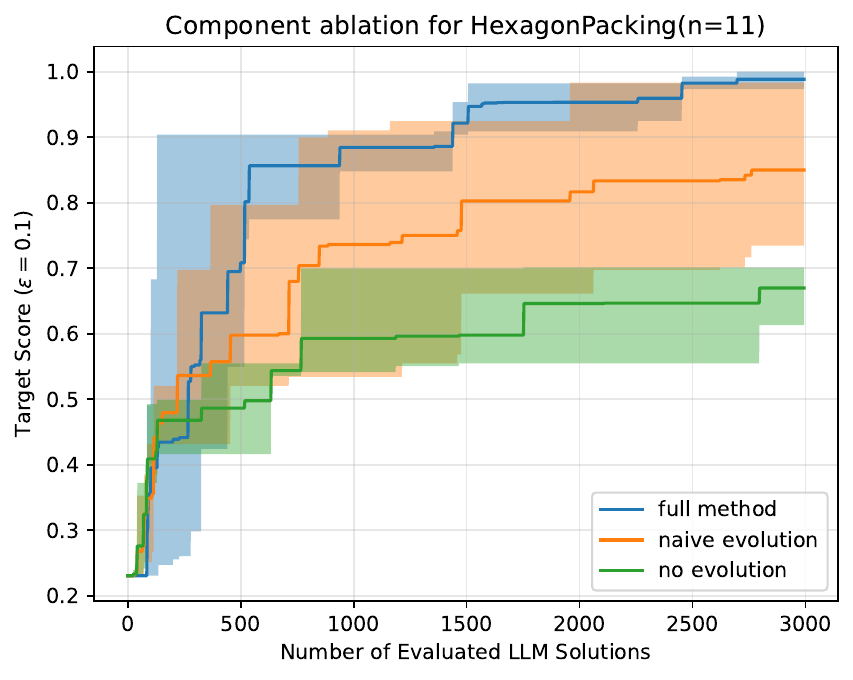} \hfill
  \includegraphics[width=0.48\linewidth]{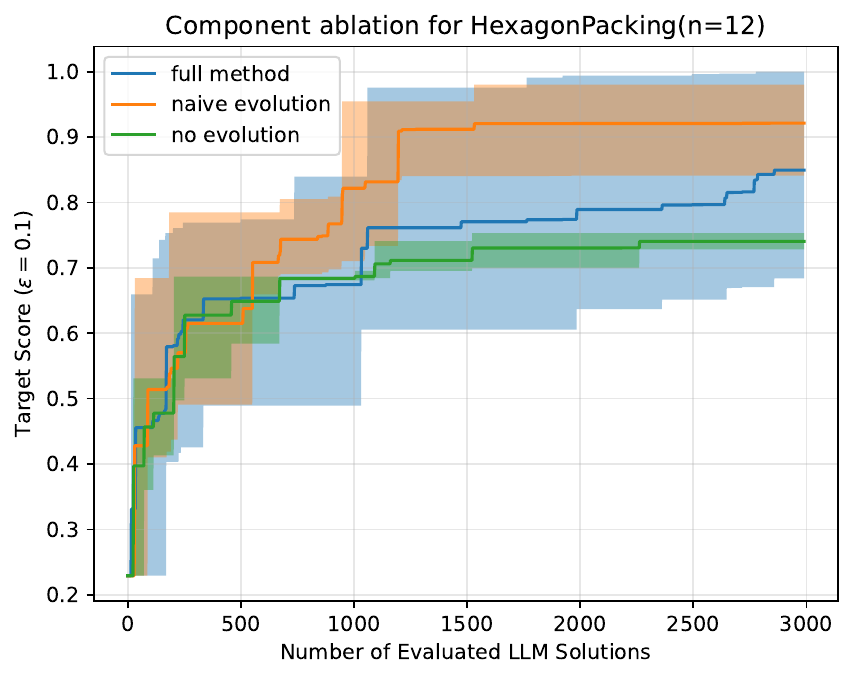}
  \caption {Component ablations for \method{} using \texttt{Qwen3-Coder-30B} on the \texttt{HexagonPacking} problem.}
  \label{fig:comp_ablation_hexpacking}
\end{figure*}

\begin{figure*}[tb]
\centering
  \includegraphics[width=0.48\linewidth]{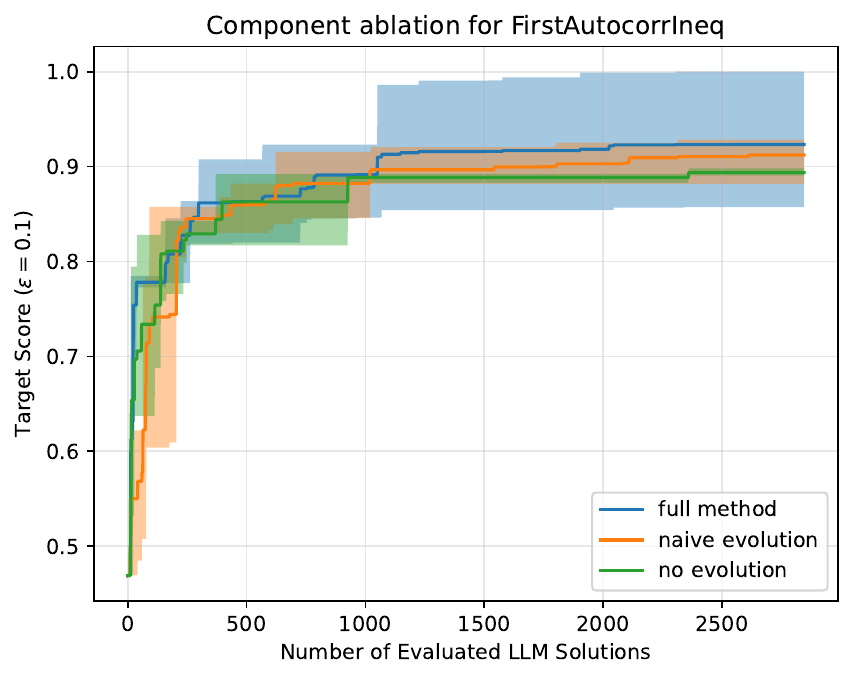} \hfill
  \includegraphics[width=0.48\linewidth]{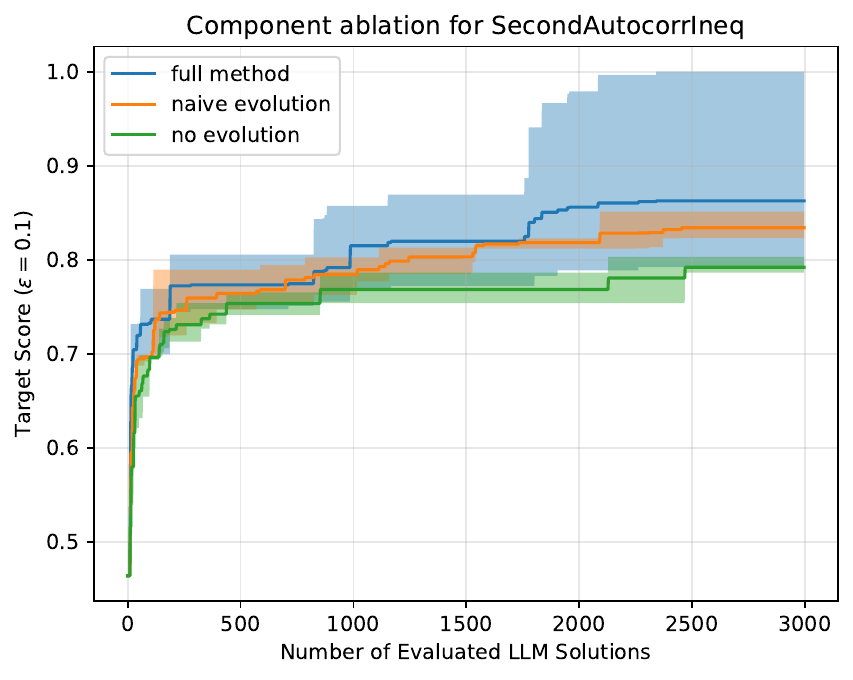}
  \caption {Component ablations for \method{} using \texttt{Qwen3-Coder-30B} on the \texttt{AutocorrIneq} problems.}
  \label{fig:comp_ablation_autocorr1}
\end{figure*}

\begin{figure*}[tb]
\centering
  \includegraphics[width=0.48\linewidth]{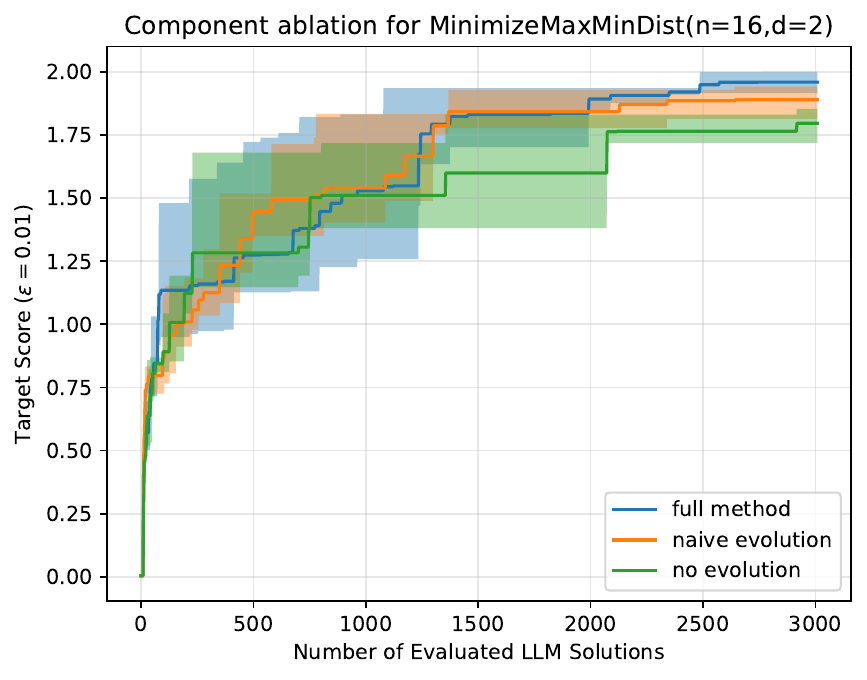} \hfill
  \includegraphics[width=0.48\linewidth]{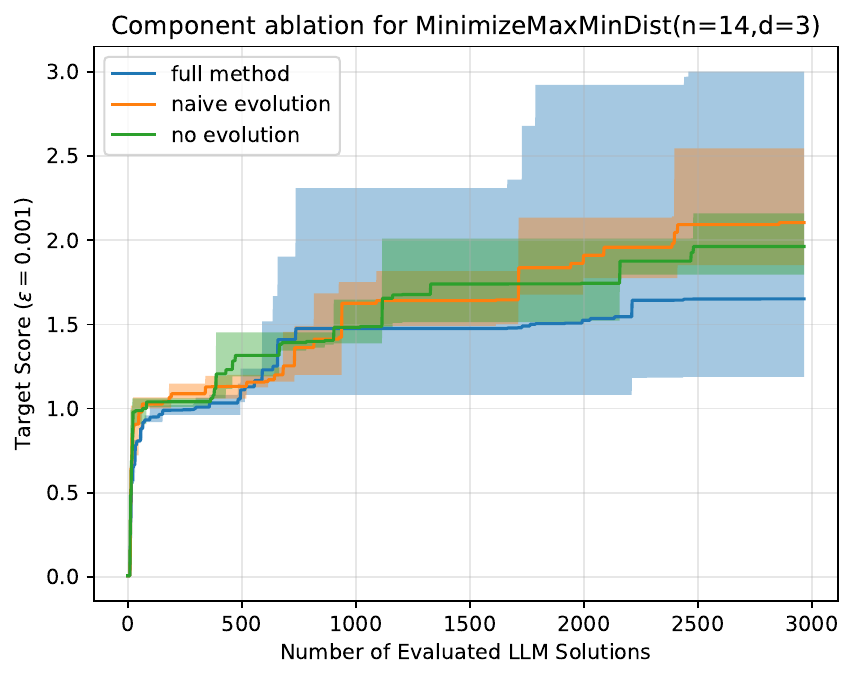}
  \caption {Component ablations for \method{} using \texttt{Qwen3-Coder-30B} on the \texttt{MinimizeMaxMinDist} problem.}
  \label{fig:comp_ablation_autocorr2}
\end{figure*}

\begin{figure}[tb]
  \includegraphics[width=0.95\columnwidth]{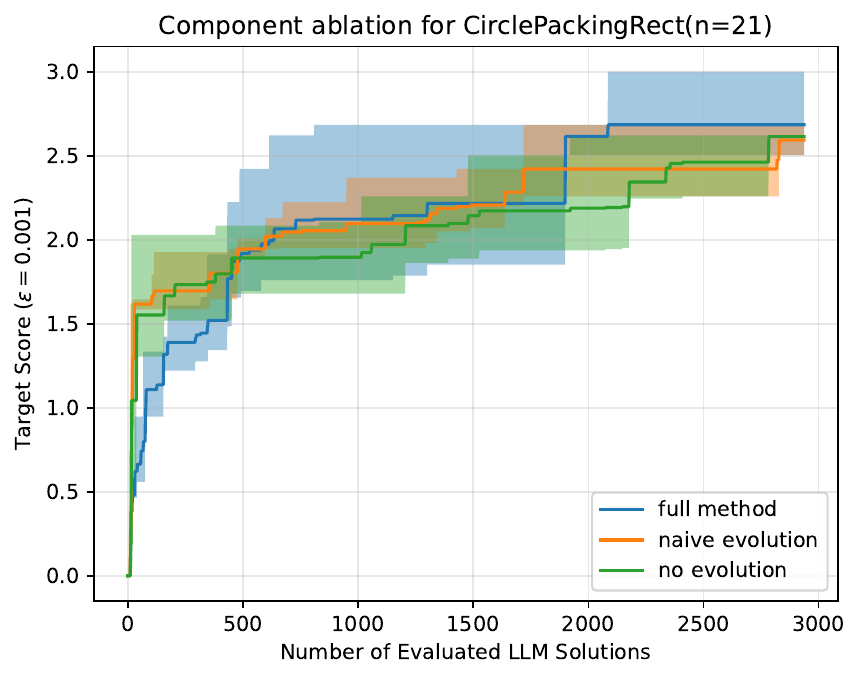}
  \caption {Component ablations for \method{} using \texttt{Qwen3-Coder-30B} on the \texttt{CirclePackingRect} problem.}
  \label{fig:comp_ablation_circle_packing_rect}
\end{figure}

\paragraph{Component ablations for remaining problems.}
Figures~\ref{fig:comp_ablation_hexpacking}--\ref{fig:comp_ablation_circle_packing_rect} display the component ablations for the remaining benchmark problems not included in Section~\ref{sec:experiments}. For the \texttt{HexagonPacking}($n=11$), \texttt{FirstAutocorrIneq}, \texttt{SecondAutocorrIneq}, \minmax{}($n=16,d=2$) and \texttt{CirclePackingRect}($n=21$) problems, the full method configuration of \method{} obtains the superior average and best results across the three runs. For the \texttt{HexagonPacking}($n=12$) and \minmax{}($n=14,d=3$) problems, although the full method does not present the highest average fitness, it also presents the best solution overall. This further reinforces the positive synergistic effect observed between \method{}'s components in Section~\ref{sec:experiments}, and shows that the full method is the key for producing the best results.

\paragraph{Impact of MAP-Elites.}
\begin{figure}[tb]
  \includegraphics[width=0.95\columnwidth]{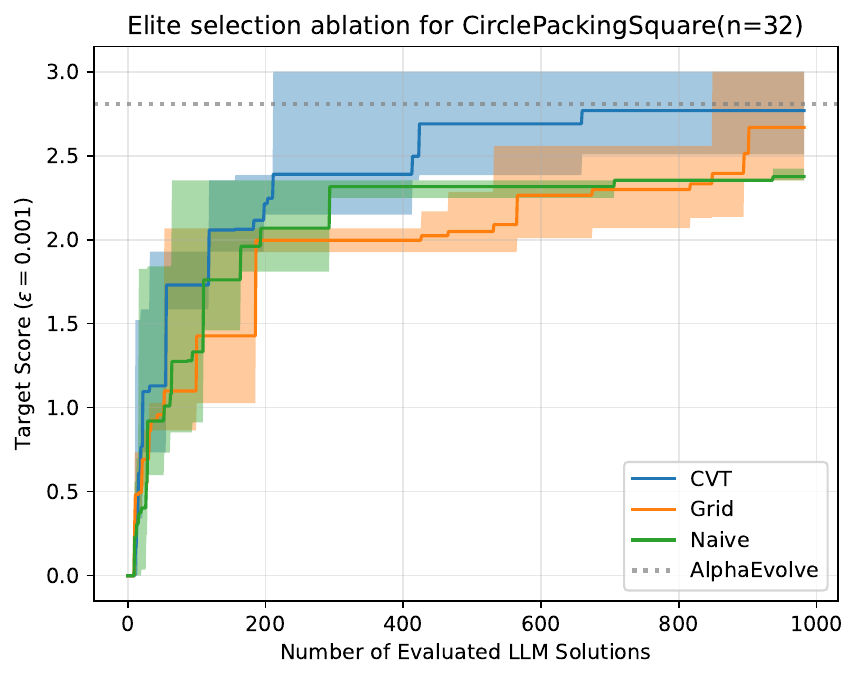} \hfill
  \caption {Elite selection ablations for \method{} using \texttt{Qwen3-Coder-30B} on the \circlepacking{} problem with $n = 32$.}
  \label{fig:appendix_ablations_elites}
\end{figure}
In this experiment, we evaluate the performance of \method{} using three distinct elite selection policies: (i) the centroidal Voronoi tessellations MAP-Elites method~\cite{vassiliades2017usingcentroidalvoronoitessellations}, referred to as \texttt{CVT}, is a variant of the MAP-Elites~\cite{mouret2015illuminatingsearchspacesmapping} algorithm, in which the feature space is partitioned according to a fixed number of centroids that approximate a centroidal Voronoi tessellation, e.g., by means of Lloyd's algorithm~\cite{1056489}; (ii) the traditional MAP-Elites method, referred to as \texttt{Grid}, in which the feature space is partitioned according to a regular lattice; and (iii) a naive elite selection, referred to as \texttt{Naive}, in which we define a maximum population cap and only add a new solution/prompt if its fitness is greater than the fitness of the worst live individual.

Figure~\ref{fig:appendix_ablations_elites} shows the results of this experiment for the \circlepacking{} problem with $n = 32$. The most notable finding is that the MAP-Elites method is clearly necessary for surpassing AlphaEvolve's results, and moreover, the \texttt{CVT} variant displays the best performance both in terms of sample efficiency and mean score. 

\paragraph{Choice of island topology.}
\begin{figure}[tb]
  \includegraphics[width=0.95\columnwidth]{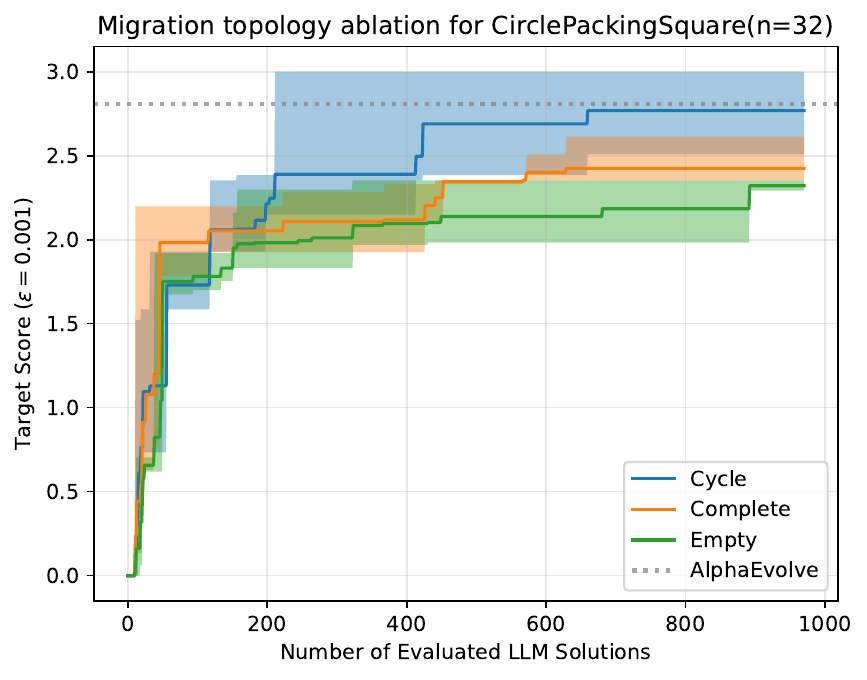} \hfill
  \caption {Migration topology ablations for \method{} using \texttt{Qwen3-Coder-30B} on the \circlepacking{} problem with $n = 32$.}
  \label{fig:appendix_ablations_top}
\end{figure}
In this experiment, we vary the underlying migration topology in order to assess its impact on the performance of \method{}. We consider three distinct topologies: (i) the \texttt{Cycle} topology connects the $N$ islands according to the undirected cycle graph $C_N$, i.e., if $\{0,...,N-1\}$ are the island indices, then island $i$ can send and receive migrants from islands $i-1\mod N$ and $i+1 \mod N$; (ii) the \texttt{Complete} topology connects the islands according to the undirected complete graph $K_N$, i.e., all islands can send and receive migrants from one another; and (iii) the \texttt{Empty} topology does not connect any of the islands, thus suppressing migration.

Figure~\ref{fig:appendix_ablations_top} shows that, for the \circlepacking{} problem with $n = 32$, the \texttt{Cycle} configuration is the only one able to surpass AlphaEvolve's results, with the \texttt{Complete} topology only being slightly superior to the \texttt{Empty} topology in mean score. This shows that migration between islands clearly plays a crucial role in producing state-of-the-art results, but also that excessively migrating between islands can be detrimental, as the overall diversity tends to decrease.

\subsection{Analysis of Search Behavior}\label{appendix:search_behavior}
Aggregate benchmark scores and component ablations establish whether the framework performs well, but do not fully explain how the search produces those results. We therefore examine four complementary questions:
\begin{enumerate}[noitemsep,label=\textbf{AQ\arabic*}]
    \item Can the operators generate executable improvements efficiently?
    \item Does diff-based editing permit substantial restructuring without uncontrolled code growth?
    \item Do evolved prompts remain useful across solution lineages and play a distinct role from exploitation?
    \item Is the observed performance stable across independent runs?
\end{enumerate}
These questions are important because the framework's practical value depends not only on final solution quality, but also on whether its search mechanisms are productive, complementary, and robust.

We analyze three representative problems covering packing, point placement, and autocorrelation: \circlepacking{} with $n=32$, \minmax{} with $n=16,d=2$, and \texttt{SecondAutocorrIneq}. Each analysis aggregates three independent runs with ten islands using \texttt{Qwen3-Coder-30B} as the single LLM backbone.

\paragraph{Operator efficiency.}
Table~\ref{tab:operator_efficiency} reports generation and evaluation outcomes separately for depth exploitation (E) and meta-prompting exploration (X). Diffs apply successfully in more than $90\%$ of attempts, while $15.5$--$27.7\%$ of all exploitation evaluations yield a strict improvement over an executable parent. Exploration is also productive, although its useful-improvement rate varies substantially by problem. Thus, the search is not dominated by malformed edits or brute-force filtering. These results answer AQ1 positively for the three representative problems, while showing that efficiency remains problem-dependent. Raw improvement rates should not be compared directly across operators: exploitation uses rank-based selection of strong parents, whereas exploration samples parents uniformly. We control for parent quality under AQ3.

\begin{table*}[tb]
\centering
\small
\caption{Operator-level statistics aggregated across three runs. E and X denote depth exploitation and meta-prompting exploration, respectively. ``Improved'' is conditioned on an executable parent; ``useful/eval'' is the number of such improvements divided by all evaluated candidates.}
\label{tab:operator_efficiency}
\resizebox{\textwidth}{!}{
\begin{tabular}{l|rr|rr|rr}
\toprule
\multirow{2}{*}{\textbf{Metric}} &
\multicolumn{2}{c|}{\circlepacking{} ($n=32$)} &
\multicolumn{2}{c|}{\minmax{} ($n=16,d=2$)} &
\multicolumn{2}{c}{\texttt{SecondAutocorrIneq}} \\
& \textbf{E} & \textbf{X} & \textbf{E} & \textbf{X} & \textbf{E} & \textbf{X} \\
\midrule
Evaluated programs & 1,663 & 598 & 4,910 & 3,414 & 4,753 & 3,418 \\
Diff-apply success & 91.5\% & 97.9\% & 95.0\% & 95.8\% & 96.0\% & 94.2\% \\
Executable rate & 66.5\% & 31.6\% & 83.1\% & 87.8\% & 70.5\% & 75.3\% \\
Improved over parent & 41.7\% & 26.4\% & 30.8\% & 23.0\% & 22.0\% & 35.4\% \\
Useful improvements/eval & 27.7\% & 3.8\% & 25.6\% & 17.2\% & 15.5\% & 19.3\% \\
Meta-prompt apply success & -- & 93.7\% & -- & 93.6\% & -- & 90.5\% \\
\bottomrule
\end{tabular}
}
\end{table*}

\paragraph{Edit granularity and program complexity.}
The \texttt{SEARCH/REPLACE} representation permits both local edits and replacement of the entire mutable \texttt{EVOLVE-BLOCK}. As Table~\ref{tab:edit_prompt_analysis} shows, \texttt{Qwen3-Coder-30B} typically selects large replacements: $86.1$--$93.8\%$ of edits span at least $80\%$ of the parent. The operator is therefore not restricted to additive local patches, although determining whether a rewrite changes the underlying algorithmic strategy remains a semantic judgment. Median growth is modest ($+4$ to $+16$ lines), and $35.6$--$45.2\%$ of accepted edits shrink or preserve program size. Fitness is uncorrelated with code length on \circlepacking{} ($n=32$) and has a modest positive correlation on the harder \minmax{} ($n=16,d=2$) and \texttt{SecondAutocorrIneq} tasks. This evidence supports a positive answer to AQ2: edits can substantially restructure the parent without consistent evidence of uncontrolled bloat. Nevertheless, the longer best solutions on the latter tasks motivate adding an explicit parsimony descriptor to MAP-Elites in future work.

\begin{table*}[tb]
\centering
\small
\caption{Edit granularity, complexity, and prompt-transfer statistics. LOC counts non-blank lines in the mutable \texttt{EVOLVE-BLOCK}; $\rho$ is Spearman's rank correlation.}
\label{tab:edit_prompt_analysis}
\resizebox{\textwidth}{!}{
\begin{tabular}{l|ccc}
\toprule
\textbf{Metric} & \circlepacking{} ($n=32$) & \minmax{} ($n=16,d=2$) & \texttt{SecondAutocorrIneq} \\
\midrule
Edits spanning $\geq 80\%$ of parent & 93.1\% & 93.8\% & 86.1\% \\
Edits touching $<15$ lines & 0.8\% & 0.9\% & 2.0\% \\
Median LOC change (child $-$ parent) & $+15$ & $+4$ & $+16$ \\
Accepted edits with non-positive LOC change & 35.6\% & 45.2\% & 37.1\% \\
$\rho$(fitness, LOC) & $-0.09$ & $0.17$ & $0.33$ \\
$\rho$(fitness, evaluation time) & $0.36$ & $0.56$ & $0.45$ \\
Mean best-solution LOC / population median LOC & 167 / 181 & 215 / 146 & 250 / 186 \\
\midrule
Prompts reused on $\geq 2$ distinct parents & 405 & 1,192 & 1,256 \\
Mean distinct lineages per reused prompt & 2.37 & 3.02 & 2.91 \\
Improvement rate, first $\rightarrow$ later uses & 50.3\% $\rightarrow$ 41.5\% & 32.9\% $\rightarrow$ 30.1\% & 41.4\% $\rightarrow$ 22.7\% \\
\bottomrule
\end{tabular}
}
\end{table*}

\paragraph{Prompt transfer and operator complementarity.}
Reused prompts span, on average, $2.4$--$3.0$ distinct solution lineages and continue to improve later parents in $22.7$--$41.5\%$ of applications (Table~\ref{tab:edit_prompt_analysis}). Prompt reuse therefore transfers useful guidance across code states, but the decrease from first to later uses---especially on \texttt{SecondAutocorrIneq}---also shows diminishing returns. Detecting redundant or harmful prompt components, and ablating them directly, remains an open direction.

To remove the parent-selection confound in the aggregate operator rates, Table~\ref{tab:operator_parent_quality} stratifies executable parents by fitness quartile. Exploration improves $42$--$59\%$ of candidates in the weakest quartile, showing that it contributes useful variation among lower-quality states. Exploitation is more reliable overall and remains productive on the fittest parents, where exploration rarely improves the parent. This division of labor is consistent with the component ablations: meta-prompting broadens the search over weaker states, while depth exploitation refines strong elites. Together, the transfer and parent-stratified results answer AQ3 positively: evolved prompts remain useful across lineages, and exploration has a distinct role from exploitation.

\begin{table*}[tb]
\centering
\small
\caption{Improvement rate over an executable parent, stratified by parent-fitness quartile. Q1 contains the weakest parents and Q4 the fittest. E and X denote exploitation and exploration.}
\label{tab:operator_parent_quality}
\resizebox{\textwidth}{!}{
\begin{tabular}{l|rr|rr|rr}
\toprule
\multirow{2}{*}{\textbf{Parent bin}} &
\multicolumn{2}{c|}{\circlepacking{} ($n=32$)} &
\multicolumn{2}{c|}{\minmax{} ($n=16,d=2$)} &
\multicolumn{2}{c}{\texttt{SecondAutocorrIneq}} \\
& \textbf{E} & \textbf{X} & \textbf{E} & \textbf{X} & \textbf{E} & \textbf{X} \\
\midrule
Q1 (weakest) & 74\% & 42\% & 68\% & 44\% & 64\% & 59\% \\
Q4 (fittest) & 13\% & 0\% & 14\% & 0\% & 12\% & 3\% \\
\bottomrule
\end{tabular}
}
\end{table*}

\paragraph{Run-to-run robustness.}
To answer AQ4, Table~\ref{tab:representative_robustness} reports the best normalized benchmark fitness from each independent run, together with the mean, $95\%$ confidence interval, and coefficient of variation. Larger values are better and $1$ denotes the reported AlphaEvolve score, allowing stability to be compared across problems with different objectives and scales.

\begin{table*}[tb]
\centering
\small
\caption{Run-to-run robustness of normalized benchmark fitness across three representative full-method runs. The confidence interval is reported as mean $\pm$ half-width.}
\label{tab:representative_robustness}
\resizebox{\textwidth}{!}{
\begin{tabular}{l|ccc}
\toprule
\textbf{Metric} & \circlepacking{} ($n=32$) & \minmax{} ($n=16,d=2$) & \texttt{SecondAutocorrIneq} \\
\midrule
Per-run best & 1.00055 / 0.99844 / 1.00055 & 0.99810 / 0.99681 / 0.99896 & 0.93673 / 0.87540 / 0.88640 \\
Mean $\pm$ 95\% CI & $0.99985 \pm 0.00303$ & $0.99795 \pm 0.00269$ & $0.89951 \pm 0.08124$ \\
Coefficient of variation & 0.122\% & 0.108\% & 3.636\% \\
\bottomrule
\end{tabular}
}
\end{table*}

The three \circlepacking{} ($n=32$) runs are tightly clustered, with a coefficient of variation of $0.122\%$; two slightly exceed the AlphaEvolve reference and the third is within $0.16\%$ of it. The \minmax{} ($n=16,d=2$) runs are similarly stable ($0.108\%$ coefficient of variation), with all three within $0.32\%$ of the reference. In contrast, \texttt{SecondAutocorrIneq} has a $3.636\%$ coefficient of variation and a substantially wider confidence interval, indicating greater sensitivity to the random seed on this harder problem. AQ4 therefore receives a positive answer for the representative packing and point-placement problems, but only a qualified answer for autocorrelation.

These confidence intervals use Student's $t$ distribution with only three runs. They summarize the observed variability but are sensitive to individual outcomes and should not be interpreted as strong evidence of statistical significance. The broader controlled comparison in Table~\ref{tab:comp_shinka_open} provides complementary context under the same three-run budget: \method{} achieves both the best solution and the best mean on six of nine problems, and its mean exceeds the best run of both OpenEvolve and ShinkaEvolve on five. Best-run performance is also a meaningful metric in algorithmic discovery because the operational objective is often to obtain at least one exceptional construction or improve a best-known bound. When the number of runs and evaluation budget are fixed and disclosed, the best result measures this discovery objective directly. Nevertheless, it can hide instability and should therefore be reported alongside per-run outcomes, means, and uncertainty, as done in Table~\ref{tab:representative_robustness}.

Overall, AQ1--AQ3 receive positive answers across all three representative problems: the operators regularly produce executable improvements, diff-based edits support broad restructuring without consistent evidence of uncontrolled bloat, and evolved prompts transfer across lineages while exploration and exploitation serve different parent-quality regimes. AQ4 is positive for the representative packing and point-placement problems but qualified for autocorrelation, where variance is higher. These diagnostics complement the endpoint and ablation results by showing that \method{}'s gains arise from productive and differentiated search mechanisms rather than only best-run selection.

\subsection{Cost and Runtime comparison}\label{appendix:cost_and_runtime}
\begin{figure*}[tb]
  \includegraphics[width=0.48\linewidth]{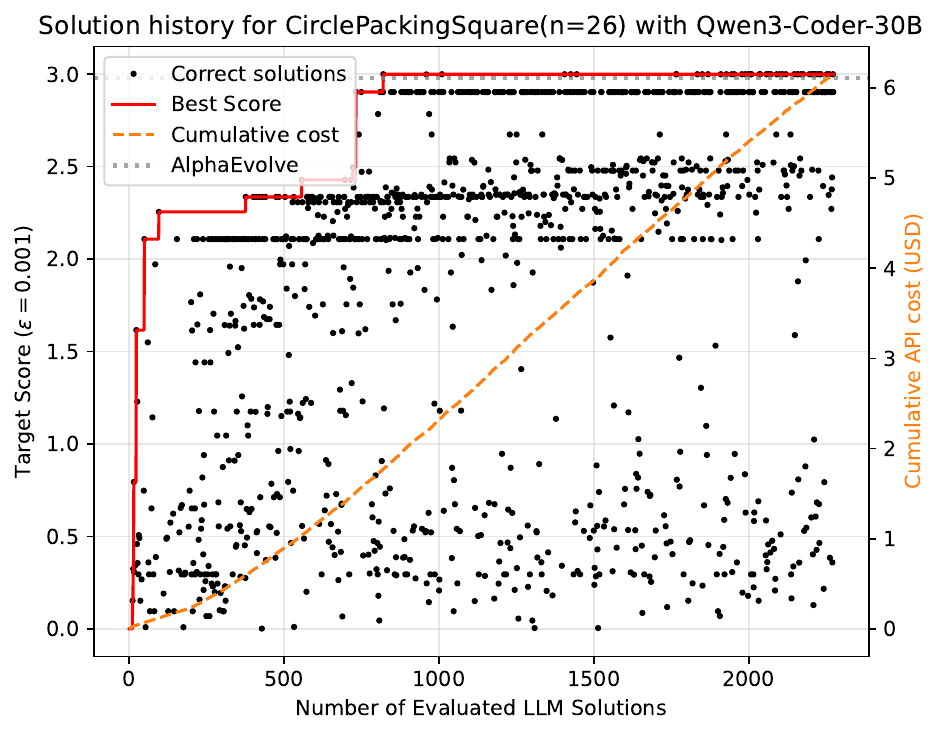} \hfill
  \includegraphics[width=0.48\linewidth]{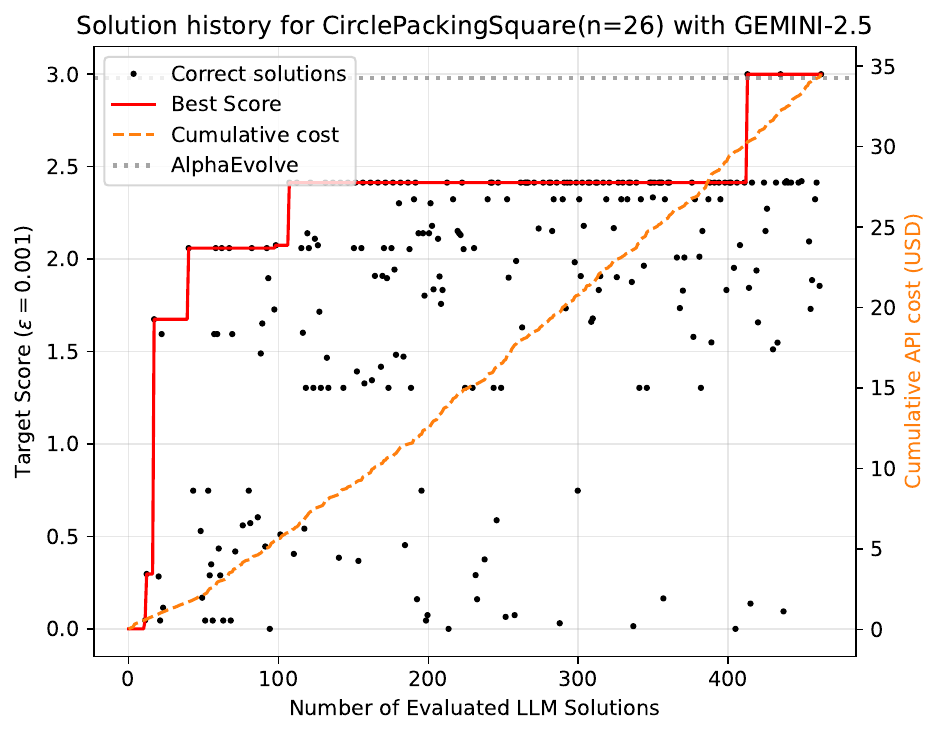}
  \caption {Solution and cost history of \texttt{Qwen3-Coder-30B} and \texttt{GEMINI-2.5} on \circlepacking{} ($n=26$). Individual points show the score of solutions executed without errors. The right vertical axis shows cumulative API cost in USD.}
  \label{fig:sol_hist_circlepackingsquare26}
\end{figure*}
\begin{figure*}[tb]
  \includegraphics[width=0.48\linewidth]{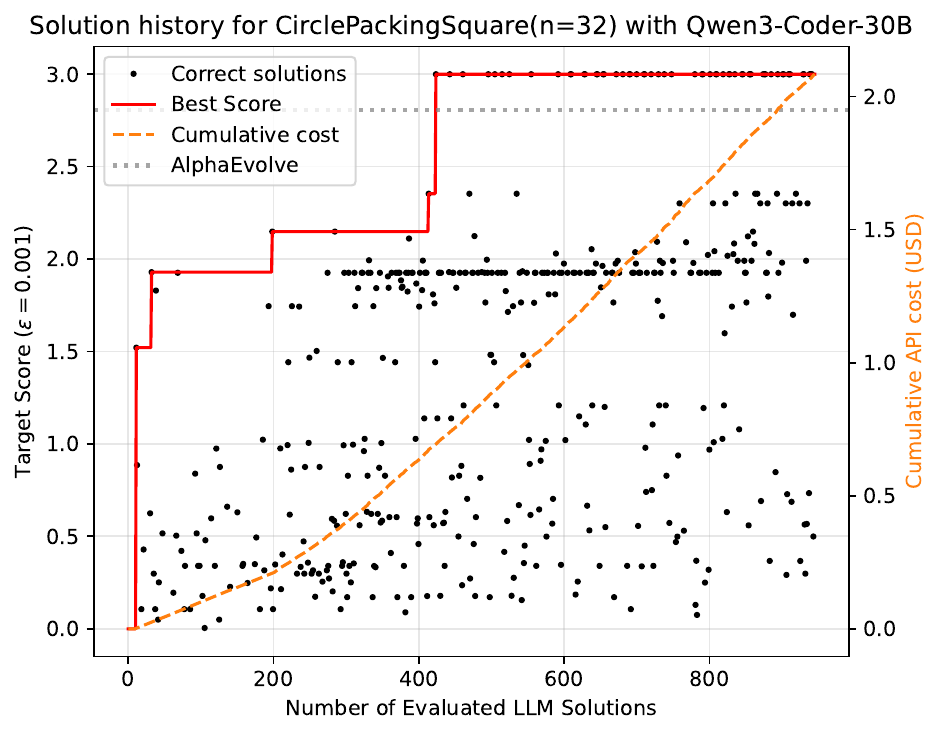} \hfill
  \includegraphics[width=0.48\linewidth]{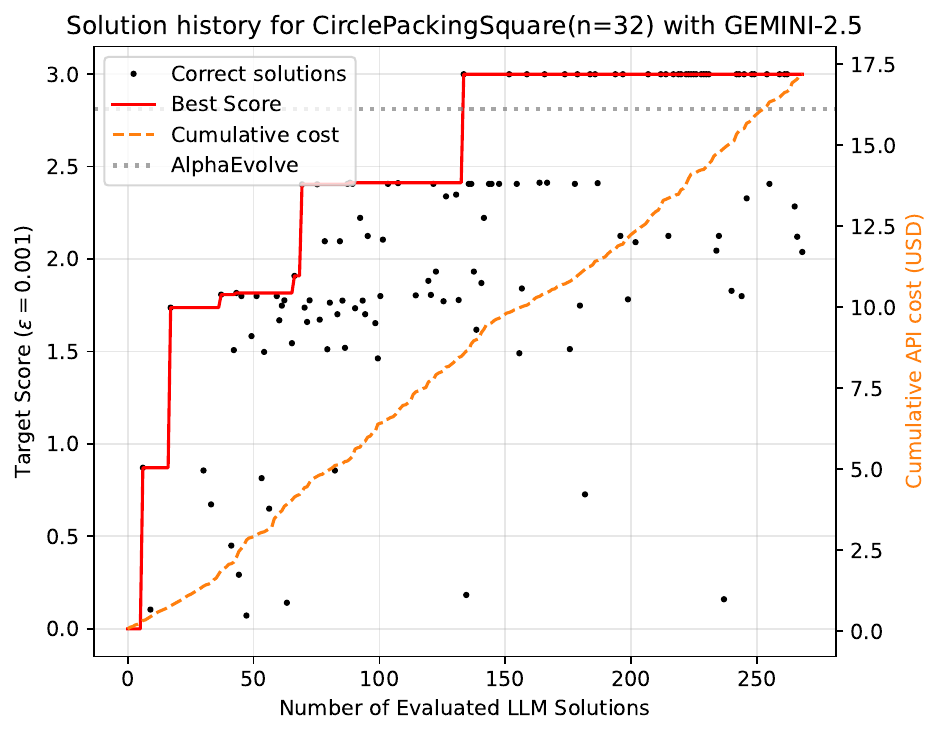}
  \caption {Solution and cost history of \texttt{Qwen3-Coder-30B} and \texttt{GEMINI-2.5} in the \circlepacking{} problem with $n=32$.}
  \label{fig:sol_hist_circlepackingsquare32}
\end{figure*}
In this section, we provide further information about the cost and runtime of our experiments. Figures~\ref{fig:sol_hist_circlepackingsquare26} and~\ref{fig:sol_hist_circlepackingsquare32} show the solution and cost histories for both ensemble configurations on the \circlepacking{} problem with $n=26$ and $n=32$, respectively. In both cases, \texttt{Qwen3-Coder-30B} achieves comparable or superior results to \texttt{GEMINI-2.5} at a fraction of the cost.

\begin{table*}[tb]
    \centering
    \small
    \caption{Cost and time comparison between \texttt{Qwen3-Coder-30B} and \texttt{GEMINI-2.5} for the best runs of \method{} (Table~\ref{tab:framework_results}) on the benchmark problems.}
    \label{tab:costs_runtime}
    \begin{tabular}{l|cc|cc}
        \toprule
        \multirow{2}{*}{\textbf{Problem}}& \multicolumn{2}{c|}{\texttt{Qwen3-Coder-30B}} & \multicolumn{2}{c}{\texttt{GEMINI-2.5}} \\
        & \textbf{Cost (USD)} & \textbf{Time (Hours)} & \textbf{Cost (USD)} & \textbf{Time (Hours)} \\
        \midrule
        
        \texttt{CirclePackingSquare($n=26$)}     & $6.2$ & $6.4$ & $34.5$ & $6.1$ \\
        \texttt{CirclePackingSquare($n=32$)}    & $2.1$ & $2.2$ & $17.2$ & $7.5$ \\
        \texttt{CirclePackingRect($n=21$)}        & $10.2$ & $12$ & $67.5$ & $11.1$ \\
        \texttt{HexagonPacking($n=11$)}             & $9.1$ & $9.8$ & $73.7$ & $17.5$ \\
        \texttt{HexagonPacking($n=12$)}           & $7.4$ & $8.5$ & $77.4$ & $15.6$ \\
        \hline
        \texttt{MinimizeMaxMinDist($n=16, d=2$)}    & $15.33$ & $25.7$ & $51.5$ & $15.7$ \\
        \texttt{MinimizeMaxMinDist($n=14, d=3$)}  & $9.6$ & $10.9$ & $54.4$ & $13.1$ \\
        \hline
        \texttt{FirstAutocorrIneq}                   & $23.1$ & $54$ & $70.8$ & $17.3$ \\
        \texttt{SecondAutocorrIneq}             & $27.2$ & $28.4$ & $66.7$ & $18.7$  \\
        
        \bottomrule
    \end{tabular}
\end{table*}

Table~\ref{tab:costs_runtime} provides approximate costs and runtimes for the best runs of both ensemble configurations on the considered benchmarks. Overall, we can easily see that \texttt{Qwen3} is significantly less expensive when compared to \texttt{GEMINI-2.5}. The runtime varies between problems, as it mainly depends on the evaluation timeout and number of islands being used (see Table~\ref{tab:codeevolve_hyper}), but overall it remains similar between configurations. For the costs and runtimes of all experiments conducted, including the multiple runs done for the ablation studies, see \codeevolvemainrepo.

\subsection{Comparison with other open-source frameworks}\label{appendix:other_os_frameworks}
Table~\ref{tab:comp_shinka_open} presents the controlled comparison of \method{} with OpenEvolve~\citep{openevolve} and ShinkaEvolve~\citep{lange2025shinka}. We adopted the same configurations and evaluation budgets (with minor adaptations to suit each framework) for all benchmark problems. Each framework was run three times with \texttt{Qwen3-Coder-30B}, using a default evaluation timeout of $60$ seconds and $10$ islands. The experimental data and results for both OpenEvolve and ShinkaEvolve are available in \codeevolvemainrepo.

\subsection{Experiments with other LLMs}\label{appendix:other_llms}

In order to evaluate \method{}'s sensitivity to the underlying LLM backbone, we test it on four backbones other than the two main ones discussed previously: \texttt{GPT-4.1}~\citep{openai2025gpt41}, \texttt{Claude Sonnet 4.5}~\citep{anthropic2025claudesonnet45}, \texttt{o4-mini}~\citep{openai2025o3o4mini} and \texttt{gpt-oss-20B}~\citep{openai2025gptoss120bgptoss20bmodel}. \texttt{GPT-4.1} and \texttt{Claude Sonnet 4.5} use the same configuration as the \texttt{GEMINI-2.5} ensemble; \texttt{o4-mini} and \texttt{gpt-oss-20B} use the same configuration as \texttt{Qwen3}. 

\begin{table}[tb]
\centering
\small
\caption{Performance of \method with different LLM backbones for the \circlepacking{} problem with $n = 32$ across three independent runs.}
\label{tab:backbone_performance}
\begin{tabular}{l|c|c}
\toprule
\textbf{Model} & \textbf{Mean} & \textbf{Best} \\
\midrule
\texttt{GPT-4.1} & 2.93685 & \textbf{2.93957} \\
\texttt{Claude Sonnet 4.5} & \textbf{2.93856} & \textbf{2.93957} \\
\midrule
\texttt{O4-MINI} & 2.93767 & \textbf{2.93957} \\
\texttt{GPT-OSS-20B} & 2.93145 & 2.93535 \\
\bottomrule
\end{tabular}
\end{table}

Table~\ref{tab:backbone_performance} reports mean and best performance across three independent runs. \texttt{GPT-4.1}, \texttt{Claude Sonnet 4.5}, and \texttt{O4-MINI} all surpass AlphaEvolve's reported best (2.93794) under the default budget without any retuning. \texttt{GPT-OSS-20B} falls short within the default budget, but surpasses AlphaEvolve when given 50 additional epochs (best: 2.93957). These results confirm that \method{} generalizes across LLM backbones without requiring retuning of prompts or operator configurations.

\section{Experiment Details}\label{appendix:configs}
In this section, we provide further details about the configurations used in our experiments. \method{}'s components have many different parameters, so we only list the most important ones here.
\subsection{Experiment Configurations}\label{appendix:experiment_configs}
\begin{table*}[tb]
    \centering
    \caption{Hyperparameters used in the best runs of \method{} (Table~\ref{tab:framework_results}) for the proposed benchmarks.}
    \label{tab:codeevolve_hyper}
    \resizebox{\textwidth}{!}{
    \begin{tabular}{l|ccccccccc}
        \toprule
        \multirow{2}{*}{\textbf{Problem}} &\multicolumn{8}{c}{\textbf{Hyperparameters}}  \\ 
        & \multirow{2}{*}{Model} & \multirow{2}{*}{Islands} & \multirow{2}{*}{Epochs} & \multirow{2}{*}{Depth} & \multirow{2}{*}{Inspirations} & \multirow{2}{*}{CPUs} & Evaluation (s) & Evaluation \\
        &  &  &  &  &  & & Timeout (s) & Max Memory (GB) \\
        \midrule
        
        \texttt{CirclePackingSquare($n=26$)} &  \texttt{Qwen3-Coder-30B} & $10$ & $250$ & $5$ & $2$ & $10$ & $60$ & $1$ \\
        \texttt{CirclePackingSquare($n=32$)} &  \texttt{Qwen3-Coder-30B} & $10$ & $100$ & $5$ & $2$ & $10$ &  $60$ & $1$ \\
        \texttt{CirclePackingRect($n=21$)}   &  \texttt{GEMINI-2.5} & $5$ & $200$ & $5$ & $2$ & $10$ & $60$ & $1$ \\
        \texttt{HexagonPacking($n=11$)}      &  \texttt{GEMINI-2.5} & $5$ & $200$ & Unlimited & $3$ & $10$ & $180$ & $5$ \\
        \texttt{HexagonPacking($n=12$)}      &  \texttt{GEMINI-2.5} & $5$ & $200$ & Unlimited & $3$ & $10$ & $180$ & $5$ \\
        \hline
        \texttt{MinimizeMaxMinDist($n=16, d=2$)}    &  \texttt{GEMINI-2.5} & $5$ & $200$ & Unlimited & $3$ & $10$ & $180$ & $5$ \\
        \texttt{MinimizeMaxMinDist($n=14, d=3$)}    &  \texttt{GEMINI-2.5} & $10$ & $100$ & Unlimited & $3$ & $20$ & $360$ & $5$ \\
        \hline
        \texttt{FirstAutocorrIneq}   &  \texttt{Qwen3-Coder-30B} & $10$ & $500$ & $5$ & $3$ & $10$ & $360$ & $5$ \\
        \texttt{SecondAutocorrIneq}  &  \texttt{Qwen3-Coder-30B} & $10$ & $900$ & $5$ & $3$ & $10$ & $60$ & $1$  \\
        
        \bottomrule
    \end{tabular}
    }
\end{table*}

All experiments with the \texttt{Qwen3-Coder-30B} model use a temperature of $0.7$ and top-p of $0.8$. The ensemble with \texttt{GEMINI-2.5 FLASH/PRO} uses temperatures of $0.7$ and top-p of $0.95$ for both models. During exploration steps, we only call the \texttt{FLASH} variant, and during exploitation steps, we call \texttt{FLASH} with $60\%$ probability and \texttt{PRO} with $40\%$ probability by default.

Table~\ref{tab:codeevolve_hyper} shows the hyperparameters for the best runs of \method{} on the proposed benchmarks (Table~\ref{tab:framework_results}). By default, we start with an exploration probability of $0.2$, and use the Plateau Scheduler to increase this probability by a multiplicative factor of $1.05$ if no fitness increase is observed for $5$ epochs, and decrease it by $0.95$ otherwise, preserving a minimum rate of $0.2$ and a maximum rate of $0.5$. 

\subsection{Hyperparameter Sensitivity and Practical Guidelines}\label{appendix:hyperparam_sensitivity}

\method{} introduces several hyperparameters. Below we classify them into three groups and provide practical guidance for new users.

\paragraph{Budget-bound parameters.}
The number of islands, epochs, evaluation timeout, and memory limit directly trade off solution quality against computational cost. Increasing any of these improves results at higher cost. We found the following defaults to be robust across all tested benchmarks: \textbf{5 islands / 200 epochs} for expensive closed-source ensembles (e.g., \texttt{GEMINI-2.5}), and \textbf{10 islands / 250 epochs} for open-weight models (e.g., \texttt{Qwen3-Coder-30B}). Evaluation timeout and memory limits are problem-specific and should be set based on the expected runtime of a single candidate solution evaluation.

\paragraph{Operator parameters.}
The maximum ancestor depth $k$ and the number of inspirations $\iota$ are the most sensitive operator-level parameters. Across all tested models and benchmarks, we found $\iota = 2$--$3$ inspirations and $k = 3$--$5$ ancestor depth to work well without problem-specific tuning. These values generalized across different backbone models (see Appendix~\ref{appendix:other_llms}).

\paragraph{Robust defaults.}
The following settings proved stable across all benchmarks and LLM backbones and serve as safe starting points: migration rate $0.1$, initial exploration rate $0.2$ with the Plateau Scheduler, ring (\texttt{Cycle}) topology, and CVT-MAP-Elites with fitness and evaluation time as feature descriptors. Task-specific meta-prompt customization is optional; a generic template is sufficient to obtain competitive results on new tasks. Complete default configurations are available at \codeevolvemainrepo.

\section{Examples of prompts and solutions}\label{appendix:prompts_sols}
In this section, we show examples of the initial prompt/solutions and the final best prompt/solution found by \method{} for the \texttt{CirclePackingSquare} problem with $n = 26$, using \texttt{Qwen3-Coder-30B}. For further information about the prompts/solutions, we refer to \codeevolvemainrepo.

\textbf{Initial prompt:}
\begin{minted}[
    linenos=false,
    breaklines,
    fontsize=\small
]{text}
  # PROMPT-BLOCK-START

  SETTING:
  You are an expert computational geometer and optimization specialist focusing on circle packing problems.
  Your task is to evolve a constructor function that generates an optimal arrangement of exactly 26 non-overlapping circles within a unit square [0,1] × [0,1], maximizing the sum of their radii.

  PROBLEM CONTEXT:
  - Target: Beat the AlphaEvolve benchmark of sum_radii = 2.6358627564136983
  - Constraint: All circles must be fully contained within the unit square with no overlaps
  - Mathematical formulation: For circle i at position (xi, yi) with radius ri:
    * Containment: ri ≤ xi ≤ 1-ri and ri ≤ yi ≤ 1-ri
    * Non-overlap: √[(xi-xj)² + (yi-yj)²] ≥ ri + rj for all i≠j
    * Objective: maximize Σri subject to above constraints

  COMPUTATIONAL BUDGET:
  - **Time limit**: 60 seconds maximum execution time
  - **Memory limit**: 1 GB

  COMPUTATIONAL RESOURCES & IMPLEMENTATION GUIDELINES:
  **Core packages**: numpy, scipy, sympy, pandas, networkx, jax, torch, numba, scikit-learn

  **Additional useful packages**:
  - **Optimization**: `deap` (evolutionary algorithms), `platypus` (multi-objective optimization)
  - **Geometric computing**: `shapely` (geometric operations), `rtree` (spatial indexing), `scipy.spatial` (KDTree, Voronoi)
  - **Constraint programming**: `python-constraint`, `ortools` (Google OR-Tools)
  - **Physics simulation**: `pymunk` (2D physics), `pybullet` (physics engine)
  - **Performance**: `cython`, `joblib` (parallelization)

  PERFORMANCE METRICS:
  1. **sum_radii**: Total sum of all 26 circle radii (PRIMARY OBJECTIVE - maximize)
  2. **benchmark_ratio**: sum_radii / 2.6358627564136983 (progress toward beating benchmark)  
  3. **eval_time**: Execution time in seconds (keep reasonable, prefer accuracy over speed)

  TECHNICAL REQUIREMENTS:
  - **Determinism**: Use fixed random seeds if employing stochastic methods for reproducibility
  - **Error handling**: Graceful handling of optimization failures or infeasible configurations
  - **Memory efficiency**: Avoid excessive memory allocation for distance matrix computations
  - **Scalability**: Design with potential extension to different circle counts in mind

  # PROMPT-BLOCK-END
\end{minted}

\textbf{Best prompt:}
\begin{minted}[
    linenos=false,
    breaklines,
    fontsize=\small
]{text}
# PROMPT-BLOCK-START

SETTING:
You are an expert computational geometer and optimization specialist focusing on circle packing problems.
Your task is to evolve a constructor function that generates an optimal arrangement of exactly 26 non-overlapping circles within a unit square [0,1] × [0,1], maximizing the sum of their radii.
Consider multiple algorithmic paradigms with emphasis on **efficient**, **fast-converging**, and **scalable** approaches that can meet the 60-second time constraint:
- **Fast heuristic methods**: Greedy placement algorithms, simulated annealing with smart cooling schedules, or local search with adaptive neighborhood structures
- **Geometric construction heuristics**: Voronoi diagrams, Delaunay triangulation, or systematic lattice arrangements with dynamic refinement
- **Discrete optimization**: Integer programming formulations, constraint satisfaction with intelligent pruning, or branch-and-bound methods
- **Hybrid approaches**: Combine fast geometric initialization with lightweight optimization (e.g., coordinate descent, alternating optimization)
- **Specialized circle packing algorithms**: Use known patterns like hexagonal close packing, or adaptations of proven packing strategies
- **Parallel computing**: Implement parallel evaluation of candidate solutions or multi-start strategies
- **Approximation algorithms**: Develop provably good approximations that converge quickly rather than exact solutions
- **Preprocessing and warm-starting**: Exploit symmetries, known optimal patterns, or previous solution information to accelerate convergence

The key is to prioritize **computational efficiency** over theoretical optimality, ensuring the solution can be computed within the 60-second time limit while still competing with the benchmark of 2.6358627564136983.

PROBLEM CONTEXT:
- Target: Beat the AlphaEvolve benchmark of sum_radii = 2.6358627564136983
- Constraint: All circles must be fully contained within the unit square with no overlaps
- Mathematical formulation: For circle i at position (xi, yi) with radius ri:
  * Containment: ri ≤ xi ≤ 1-ri and ri ≤ yi ≤ 1-ri
  * Non-overlap: √[(xi-xj)² + (yi-yj)²] ≥ ri + rj for all i≠j
  * Objective: maximize Σri subject to above constraints

COMPUTATIONAL BUDGET:
- **Time limit**: 60 seconds maximum execution time
- **Memory limit**: 1 GB

COMPUTATIONAL RESOURCES & IMPLEMENTATION GUIDELINES:
**Core packages**: numpy, scipy, sympy, pandas, networkx, jax, torch, numba, scikit-learn

**Additional useful packages**:
- **Optimization**: `deap` (evolutionary algorithms), `platypus` (multi-objective optimization), `scipy.optimize` (numerical optimization)
- **Geometric computing**: `shapely` (geometric operations), `rtree` (spatial indexing), `scipy.spatial` (KDTree, Voronoi)
- **Constraint programming**: `python-constraint`, `ortools` (Google OR-Tools)
- **Physics simulation**: `pymunk` (2D physics), `pybullet` (physics engine)
- **Performance**: `cython`, `joblib` (parallelization)
- **Advanced techniques**: Consider using `numba.jit` for performance-critical geometric computations or `jax` for automatic differentiation in gradient-based methods
- **Specialized solvers**: Consider `cvxpy` for convex optimization, `gurobipy` for commercial solvers, or `z3` for SMT solving
- **Machine learning**: Explore `torch` for neural network approaches or `stable-baselines3` for reinforcement learning

PERFORMANCE METRICS:
1. **sum_radii**: Total sum of all 26 circle radii (PRIMARY OBJECTIVE - maximize)
2. **benchmark_ratio**: sum_radii / 2.6358627564136983 (progress toward beating benchmark)  
3. **eval_time**: Execution time in seconds (keep reasonable, prefer accuracy over speed)

TECHNICAL REQUIREMENTS:
- **Determinism**: Use fixed random seeds if employing stochastic methods for reproducibility
- **Error handling**: Graceful handling of optimization failures or infeasible configurations
- **Memory efficiency**: Avoid excessive memory allocation for distance matrix computations
- **Scalability**: Design with potential extension to different circle counts in mind
- **Robustness**: Ensure all generated solutions are mathematically valid (no overlaps, all circles contained)
- **Alternative approaches**: Prefer methods that guarantee valid starting configurations over heuristic initialization methods that may produce invalid placements

# PROMPT-BLOCK-END
    
\end{minted}

\textbf{Initial solution:}
\begin{minted}[
    linenos=false,
    breaklines,
    fontsize=\small
]{python}
# EVOLVE-BLOCK-START
import numpy as np

# You can define functions outside the main function below.
# Remember that any function used in parallel computation must be defined globally and not locally.

def circle_packing26() -> np.ndarray:
    """
    Places 26 non-overlapping circles in the unit square in order to maximize the sum of radii.

    Returns:
        circles: np.array of shape (26,3), where the i-th row (x,y,r) stores the (x,y) coordinates of the i-th circle of radius r.
    """
    n = 26
    circles = np.zeros((n, 3))

    return circles


# EVOLVE-BLOCK-END
\end{minted}

\textbf{Best solution:}
\begin{minted}[
    linenos=false,
    breaklines,
    fontsize=\small
]{python}
# EVOLVE-BLOCK-START
import numpy as np
from scipy.optimize import minimize
import math

def circle_packing26() -> np.ndarray:
    """
    Places 26 non-overlapping circles in the unit square in order to maximize the sum of radii.
    Uses a hybrid approach combining geometric initialization and constrained optimization.
    
    Returns:
        circles: np.array of shape (26,3), where the i-th row (x,y,r) stores the (x,y) coordinates 
                 of the i-th circle of radius r.
    """
    n = 26
    
    # Initialize using a better hexagonal layout that starts closer to optimal (from inspiration 2)
    def initialize_better_hexagonal():
        circles = []
        
        # Use a 5x5 hexagonal grid with optimized spacing
        rows = 5
        cols = 5
        
        # Calculate spacing more carefully to allow for better optimization
        initial_radius = 0.08  # Slightly smaller initial radius to allow expansion
        spacing_x = 2 * initial_radius
        spacing_y = initial_radius * math.sqrt(3)
        
        # Center the pattern in the unit square
        offset_x = (1.0 - (cols - 1) * spacing_x) / 2
        offset_y = (1.0 - (rows - 1) * spacing_y) / 2
        
        # Generate hexagonal pattern
        for i in range(rows):
            for j in range(cols):
                if len(circles) >= n:
                    break
                # Offset every other row
                x_offset = (i % 2) * spacing_x / 2
                x = offset_x + j * spacing_x + x_offset
                y = offset_y + i * spacing_y
                
                # Add more substantial randomness to avoid getting stuck in local minima
                x += np.random.uniform(-spacing_x/8, spacing_x/8)
                y += np.random.uniform(-spacing_y/8, spacing_y/8)
                
                # Ensure we're within bounds with safety margin
                x = max(initial_radius, min(1 - initial_radius, x))
                y = max(initial_radius, min(1 - initial_radius, y))
                
                circles.append([x, y, initial_radius])
        
        # Trim or pad to exactly 26 circles
        if len(circles) > n:
            circles = circles[:n]
        elif len(circles) < n:
            # Fill remaining slots with strategic positioning
            for i in range(n - len(circles)):
                # Place in center region with more varied positions
                x = 0.2 + 0.6 * np.random.random()
                y = 0.2 + 0.6 * np.random.random()
                r = 0.05 + 0.03 * np.random.random()  # Variable initial radii
                circles.append([x, y, r])
        
        return np.array(circles)
    
    # Vectorized constraint functions for better performance (from inspiration 2)
    def constraint_containment(circles_flat):
        """Vectorized containment constraints"""
        circles = circles_flat.reshape(-1, 3)
        x = circles[:, 0]
        y = circles[:, 1]
        r = circles[:, 2]
        
        # All four boundary constraints in one go
        return np.concatenate([
            x - r,           # Left boundary
            1 - x - r,       # Right boundary  
            y - r,           # Bottom boundary
            1 - y - r        # Top boundary
        ])
    
    def constraint_nonoverlap(circles_flat):
        """Fully vectorized non-overlap constraints"""
        circles = circles_flat.reshape(-1, 3)
        
        # Create all pairwise comparisons using broadcasting for maximum efficiency
        n_circles = len(circles)
        if n_circles < 2:
            return np.array([])
        
        # Extract coordinates and radii
        x = circles[:, 0]
        y = circles[:, 1]
        r = circles[:, 2]
        
        # Use broadcasting to compute all pairwise distances at once
        # Reshape for broadcasting: (n, 1) and (1, n)
        x_i = x.reshape(-1, 1)
        y_i = y.reshape(-1, 1)
        r_i = r.reshape(-1, 1)
        
        x_j = x.reshape(1, -1)
        y_j = y.reshape(1, -1)
        r_j = r.reshape(1, -1)
        
        # Compute squared distances between all pairs
        dist_sq = (x_i - x_j)**2 + (y_i - y_j)**2
        
        # Compute minimum required distances (sum of radii)
        min_dist_sq = (r_i + r_j)**2
        
        # We want dist_sq >= min_dist_sq, so we return dist_sq - min_dist_sq
        # But we only want the upper triangle (unique pairs) to avoid duplicates
        mask = np.triu(np.ones((n_circles, n_circles), dtype=bool), k=1)
        constraints = dist_sq[mask] - min_dist_sq[mask]
        
        # Add a small epsilon to avoid numerical issues with very tight constraints
        # This helps with convergence stability
        return constraints + 1e-12
    
    # Optimization with better convergence control
    def optimize_with_tightening(initial_params, bounds, constraints):
        """Run optimization with tighter convergence criteria for better results"""
        # Try multiple optimization runs with different tolerances
        methods = ['SLSQP', 'trust-constr']
        best_result = None
        best_sum = -float('inf')
        
        for method in methods:
            try:
                result = minimize(
                    objective,
                    initial_params,
                    method=method,
                    bounds=bounds,
                    constraints=constraints,
                    options={'maxiter': 1000, 'ftol': 1e-12, 'gtol': 1e-12, 'disp': False}
                )
                
                if result.success:
                    current_sum = -result.fun
                    if current_sum > best_sum:
                        best_sum = current_sum
                        best_result = result
            except Exception:
                continue
        
        return best_result
    
    # Multiple restarts with better strategies (from inspiration 2)
    best_result = None
    best_sum = -float('inf')
    
    # Try fewer but more strategic restarts (10 instead of 15)
    max_restarts = 10
    
    for restart in range(max_restarts):
        # Set seed for reproducibility
        np.random.seed(restart * 1000 + 42)
        
        # Initialize circles using better approach
        circles = initialize_better_hexagonal()
        
        # Define optimization bounds: [x1, y1, r1, x2, y2, r2, ...]
        bounds = []
        for i in range(n):
            # x coordinate bounds: [r, 1-r] (ensuring circle fits)
            # y coordinate bounds: [r, 1-r] (ensuring circle fits)
            # r coordinate bounds: [1e-6, 0.4] (tighter upper bound)
            bounds.extend([(1e-6, 0.999), (1e-6, 0.999), (1e-6, 0.4)])
        
        # Flatten initial circles for optimization
        initial_params = circles.flatten()
        
        # Objective function: minimize negative sum of radii (to maximize sum)
        def objective(params):
            circles_local = params.reshape(-1, 3)
            return -np.sum(circles_local[:, 2])
        
        # Set up constraints for scipy.optimize using vectorized versions
        constraints = [
            {'type': 'ineq', 'fun': constraint_containment},
            {'type': 'ineq', 'fun': constraint_nonoverlap}
        ]
        
        # Run optimization with both methods for robustness (from inspiration 2)
        methods = ['SLSQP', 'trust-constr']
        local_best_result = None
        local_best_sum = -float('inf')
        
        for method in methods:
            try:
                result = minimize(
                    objective,
                    initial_params,
                    method=method,
                    bounds=bounds,
                    constraints=constraints,
                    options={'maxiter': 1000, 'ftol': 1e-10, 'gtol': 1e-10, 'disp': False}
                )
                
                if result.success:
                    current_sum = -result.fun
                    if current_sum > local_best_sum:
                        local_best_sum = current_sum
                        local_best_result = result
            except Exception:
                continue  # Skip this method if it fails
        
        # Update global best if we found something better
        if local_best_result is not None and local_best_result.success:
            current_sum = -local_best_result.fun
            if current_sum > best_sum:
                best_sum = current_sum
                best_result = local_best_result
    
    # Return the best result found
    if best_result is not None and best_result.success:
        final_circles = best_result.x.reshape(-1, 3)
        # Ensure all circles are properly contained and valid
        for i in range(len(final_circles)):
            x, y, r = final_circles[i]
            # Clamp positions to be within safe bounds
            x = max(r, min(1-r, x))
            y = max(r, min(1-r, y))
            # Clamp radius to reasonable bounds
            r = max(1e-6, min(0.4, r))
            final_circles[i] = [x, y, r]
        return final_circles
    else:
        # If all optimizations fail, return the best initial configuration
        return initialize_better_hexagonal()


# EVOLVE-BLOCK-END

\end{minted}

\end{document}